\documentclass[10pt,twocolumn,letterpaper]{article}

\usepackage[pagenumbers]{cvpr}
\usepackage{natbib}
\usepackage{multicol,multirow,array}
\usepackage{hyperref}
\usepackage{url}
\usepackage{float}
\usepackage{amsmath,amssymb,amsfonts,bm,mathtools}
\usepackage{graphicx}
\usepackage{textcomp}
\usepackage{xcolor}
\usepackage{soul}
\usepackage{lipsum}
\usepackage{algorithm, setspace}
\usepackage{algpseudocode}
\usepackage{arydshln}
\usepackage{pifont}

\usepackage{gensymb}
\usepackage{xspace}
\usepackage{subcaption}
\usepackage{pgfmath} % or tikz
\usepackage{dblfloatfix} % Fix for two-column float placement

\usepackage[normalem]{ulem}

\usepackage{wrapfig}
\def\etc{\textit{etc.\xspace}}

\usepackage{makecell}
\long\def\invis#1{}

\usepackage[shortlabels]{enumitem}

\newcommand\hidden[1]{}
\newcommand\z{\phantom{0}}

\newcommand{\li}[1]{\noindent\textbf{#1}}

\newcommand\DOTS{\hbox to 1em{.\hss.\hss.}}
\newcommand{\Prompt}[1]{``{\tt #1}''}
\newcommand{\DscPrompt}[1]{\texttt{\small{\textbf{[#1]}}}}
\newcommand{\SttPrompt}[1]{\texttt{\small{\textbf{(#1)}}}}
\newcommand{\ObjPrompt}[1]{\texttt{\small{\textbf{\{#1\}}}}}
\newcommand{\ttt}[1]{\texttt{\small{\textbf{#1}}}}
\newcommand{\DIV}{\allowbreak |\allowbreak}

\def\ROVER{\textit{Open\,Rover}}

\def\BRAND{\textit{VL-Explore}}

\def\Ours{\textbf{\BRAND\,\ding{72}}}

\def\C{C.}
\def\NW{N.W.}
\def\NE{N.\,E.}
\def\SW{S.W.}
\def\SE{S.\,E.}

\newcommand{\SUPPLEMNTARY}[1]{#1 of the supplementary materials}
\newcommand{\EP}{\mathbf{\text{EPS}}}

\begin{document}
\title{\BRAND: Zero-shot Vision-Language Exploration and Target Discovery \\ by Mobile Robots
\vspace{-2mm}
}
\author{Yuxuan Zhang$^{1}$, Adnan Abdullah$^{2}$, Sanjeev J. Koppal$^{\star3}$, and Md Jahidul Islam$^{\star4}$\\
{\small$^{\star}$Equal Contribution}
\thanks{$^{2,4}$RoboPI Laboratory, Dept. of ECE, University of Florida (UF)}
\thanks{$^{1,3}$FOCUS Laboratory, Dept. of ECE, University of Florida (UF)}
\thanks{$^{3}$Amazon Robotics (Dr. Koppal holds concurrent appointments as an Associate Professor of ECE-UF and an Amazon Scholar at Amazon Robotics. This project is performed at UF, not associated with Amazon.)}
\vspace{-5mm}
}

%% Banner figure
% \makeatletter
% \g@addto@macro\@maketitle{
% \begin{figure}[H]
%     \centering
%     \begin{minipage}{\textwidth}
%     \vspace{-5mm}
%     \includegraphics[width=\textwidth]{res/FirstPageBanner.pdf}%
%     \vspace{-2mm}
%     \caption{Illustration of an ongoing exploration and target discovery task by the proposed ClipRover system. \textbf{Left}: A synthesized image representing the robot's onboard camera; the cone illustrates its field-of-view, divided into six numbered tiles for detailed analysis. \textbf{Right}:
%     The vision-language perception of each tile is processed by a novel correlation middleware; 
%     The motion mixer engine is currently in `target lock' mode, prioritizing the target (a teddy bear) in tile-5  over other navigable regions (\eg, tile-6) due to the higher precedence assigned to the target. Full demo is available at the project page: \url{https://robopi.ece.ufl.edu/cliprover.html}.
%     }
%     \label{fig:first-page-banner}
%     \end{minipage}
%     \vspace{-1mm}
% \end{figure}
% }
% \makeatother
\maketitle

\begin{abstract}
Vision-language navigation (VLN) has emerged as a promising paradigm, enabling mobile robots to perform zero-shot inference and execute tasks without specific pre-programming. However, current systems often separate map exploration and path planning, with exploration relying on inefficient algorithms due to limited (partially observed) environmental information. In this paper, we present a novel navigation pipeline named ``\BRAND" for simultaneous exploration and target discovery in unknown environments, leveraging the capabilities of a vision-language model named CLIP. Our approach requires only monocular vision and operates without any prior map or knowledge about the target. For comprehensive evaluations, we designed a functional prototype of a UGV (unmanned ground vehicle) system named ``\ROVER", a customized platform for general-purpose VLN tasks. We integrated and deployed the \BRAND~ pipeline on \ROVER~ to evaluate its throughput, obstacle avoidance capability, and trajectory performance across various real-world scenarios. Experimental results demonstrate that \BRAND~ consistently outperforms traditional map-traversal algorithms and achieves performance comparable to path-planning methods that depend on prior map and target knowledge. Notably, \BRAND~ offers real-time active navigation without requiring pre-captured candidate images or pre-built node graphs, addressing key limitations of existing VLN pipelines.
\end{abstract}

% \begin{IEEEkeywords}

% \IEEEpeerreviewmaketitle
\section{Introduction}
\vspace{-1mm}
Autonomous robots face considerable challenges in exploring and discovering targets within unknown environments without a prior map, typically requiring a dedicated mapping phase before initiating path planning strategies \citep{Ginting2024Seek, kiran2022spatialrelationgraphgraph, Savarese-RSS-19}. Recent vision-language (VL) approaches have begun to address this by incorporating long-term memory architectures, a key factor driving performance gains in navigation tasks~\citep{chang2023goatthing}. However, methods that rely on pre-constructed maps or pre-existing environmental knowledge often exhibit poor adaptability to dynamic or evolving conditions, necessitating frequent reinitialization or bespoke algorithmic interventions~\citep{hahnel2003dynamic, abujabal2024comprehensive}. These limitations are especially pronounced in critical real-world applications—such as search-and-rescue, environmental monitoring, and warehouse exploration—where adaptability and robustness are essential~\citep{keith2024review, wang2019autonomous, lluvia2021active}. Despite the importance of efficient, zero-shot exploration in such scenarios, current VLN systems have yet to adequately address this capability.

\begin{figure}[t]
\centering
\includegraphics[width=0.92\linewidth]{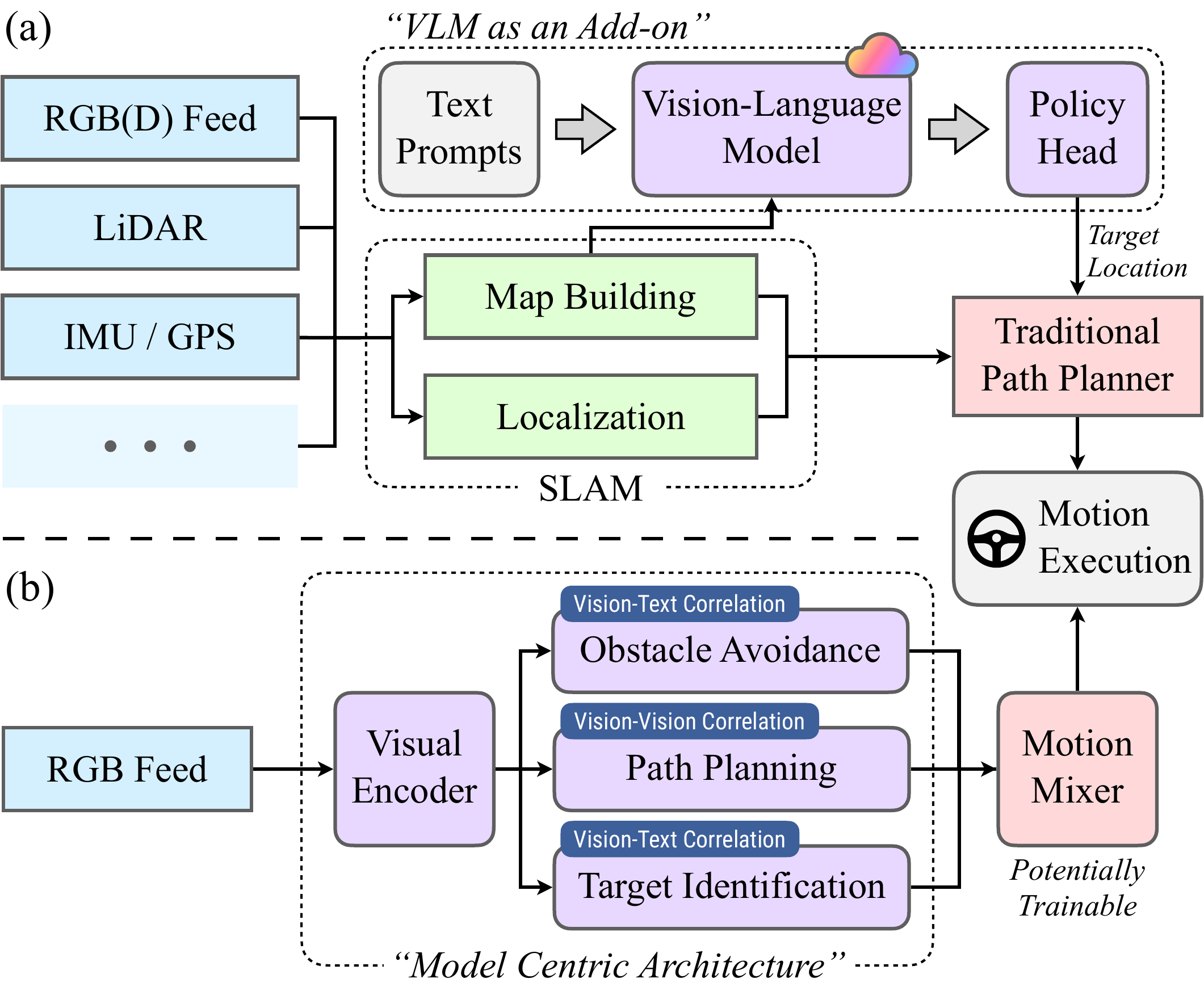}%
\vspace{-1mm}
\caption{(a) Existing VLN systems generally use VLMs as a complementary add-on;  (b) The proposed \BRAND ~pipeline is designed with a general-purpose VLM at its core, eliminating the need for multimodal sensing and extra processing steps.
%Comparison of the proposed \BRAND\allowbreak~ system (b) with existing VLN systems (a). Unlike most existing VLN systems that , 
}
\label{fig:arch-comparison}
\vspace{-2mm}
\end{figure}

\begin{figure*}[ht]
    \centering
    % \begin{minipage}{\textwidth}
    \includegraphics[width=\textwidth]{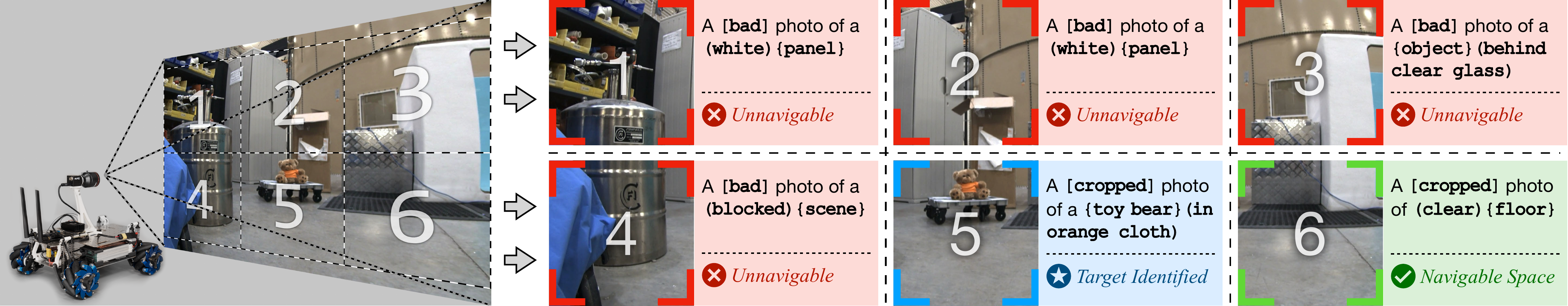}%
    \vspace{-1mm}
    \caption{Illustration of an ongoing exploration and target discovery task by the proposed \BRAND ~ system. \textbf{Left}: A synthesized image representing the robot's perspective from its onboard camera; the cone illustrates its field-of-view, divided into six numbered tiles for detailed analysis. \textbf{Right}: The vision-language perception of each tile is processed by a novel correlation middleware;
    % distinct modules generate the respective \textit{navigability scores} and \textit{target confidence values} as shown.
    The motion mixer engine is currently in `target lock' mode, prioritizing the target (a teddy bear) in tile-5  over other navigable regions (\eg, tile-6) due to the higher precedence assigned to the target. %\JI{I revised the caption; btw, the nav scores and conf values are mentioned to be there, but no values are in the image.}
    %\draft{A snapshot of an ongoing exploration \& discovery task using our proposed system. The robot has locked onto its target and is slowly approaching it. \li{Left:} A synthesized image showing the robot's view from its own camera. The cone represents the robot's field of view. The frame is sliced into 6 tiles and numbered accordingly; \li{Right:} The robot's perception of each tile, generated by correlation middlewares. The \textit{navigability scores} and \textit{target confidences} are generated by different middlewares. At this moment, the motion mixer backend is in target lock mode (\ie~approaching tile 5 while ignoring tile 6) since the target has higher precedence over normal navigable spaces.}
    }%
    \label{fig:first-page-banner}
    % \end{minipage}
    \vspace{-1mm}
\end{figure*}

% Contemporary vision-language-navigation (VLN) systems have been designed to optimize performance in already explored or static environments, often overlooking the efficiency and reliability of first-time exploration in unknown settings. %Most existing VLN systems prioritize optimizing navigation performance in, conducting mapping and target discovery as separate processes. 
When initializing inside an unexplored environment, many traditional methods use a fixed policy that often gets stuck in local minima, leading to inefficient or incomplete exploration \citep{khatib1985real}. To address this, contemporary VLN systems use high-sensing modalities \citep{Ginting2024Seek,chang2023goatthing} such as multiple cameras, depth camera or LiDAR to help with first-time exploration. For instance, a 2D LiDAR is commonly used for obstacle avoidance and mapping~\citep{sun2023autonomous,peng2015obstacle}, while a separate camera is used for scene classification and target detection~\citep{kim2022autonomous}.
Additionally, the initial map traversal is not feasible for partially observable and dynamic environments, requiring frequent re-initialization~\citep{hernandez2020real,bonnevie2021long}. To this end, the efficiency and robustness of target discovery by \textbf{zero-shot exploration} remain largely under-explored.

In this paper, we present ``\textbf{VL-Explore}", a novel navigation framework for zero-shot exploration and target discovery by unmanned ground vehicles (UGVs) in unknown environments. VL-Explore extends the spatial context awareness capabilities of general-purpose VLMs~\citep{chen2024spatialvlm} to guide 2D robotic exploration and target discovery without requiring access to a prior map. It addresses the limitation that existing VLN systems predominantly use a VLM as a complementary add-on for policy projection, outside the navigation subsystem; see Fig. \ref{fig:arch-comparison}. Instead, \textbf{\BRAND ~ uses a general-purpose VLM as its navigation core}, with the entire system operating within it. It also eliminates the need for any additional sensing modalities and map building. As illustrated in Fig. \ref{fig:arch-comparison}b, simplifying the system architecture in this way significantly reduces the system complexity and cost, making it more scalable and suitable for resource-constrained systems.

% It presents a modular architecture organized into three key stages: \textit{perception}, \textit{correlation}, and \textit{decision}. This modular design enables the decomposition of complex system-level challenges into smaller, manageable sub-problems, allowing for iterative improvements to each stage. Furthermore, the framework features a highly configurable correlation middleware, offering flexibility to adapt to various tasks and environmental conditions without any prior map or knowledge about the target. \JI{very poor paragraph: why are you talking about modular system architecture so much? Mention it, but you got to talk about what is the algorithmic (scientific) novelty of the proposed work}

% With vision-language integration - demonstrate superior navigation performance while providing semantic scene understanding. However, these systems 

VL-Explore eliminates the contemporary VLM system's \textbf{reliance on pre-built knowledge graphs} and candidate image libraries. Specifically, SOTA systems such as Clip-Nav~\citep{dorbala2022clipnav}, Clip on Wheels~\citep{Gadre2022CoWsOP}, and GCN~\citep{kiran2022spatialrelationgraphgraph} require pre-captured images of candidate scenes, target locations, and/or pre-built knowledge graphs of the navigable space as prior~\citep{Savarese-RSS-19}. Due to this complexity, they are generally tested in 2D simulations or overly simplistic environments. Other contemporary methods, such as SEEK~\citep{Ginting2024Seek}, require high-sensing modalities involving multiple cameras, LiDAR, and onboard SLAM pipelines for navigation.
In \BRAND, we formulate the problem from a different perspective. Instead of relying on a separate traditional mapping and path-planning backbone, our proposed architecture uses the visual embeddings encoded by a foundational VLM as the \textbf{only input to the system}. High-level functionalities such as obstacle avoidance, path planning, and target identification are built upon this foundation for active visual perception.

% highlight/list your contributions and impact / use cases
To validate the proposed framework in real-world environments, we develop a UGV platform that meets the computational demands of VLMs while offering superior maneuverability and mechanical stability during the task. We deployed the CLIP model~\citep{radford2021learning,ilharco_gabriel_2021_5143773} in our proposed architecture and performed extensive tests in a real-world environment. The results demonstrate that VL-Explore consistently outperforms map-traversal algorithms and achieves performance comparable to path-finding methods, despite the latter relying on prior knowledge of the map and target location. Moreover, VL-Explore achieves significantly shorter trajectory lengths, making it more efficient than map-traversal algorithms. In comparison with state-of-the-art systems, VL-Explore achieves comparable or even superior performance to systems with higher sensor modalities or even with long-term memory. Notably, VL-Explore offers such performance margins without additional sensor modalities, also requiring no separate dedicated map-building or localization modules.

% \JI{This paragraph should dive into the details of the proposed system, highlighting the novel features and impact}

\vspace{2mm}
\noindent
Overall, we make the following contributions in this paper:
\vspace{1mm}

\begin{enumerate}[label={$\arabic*$.},nolistsep,leftmargin=*]

\item We propose a model-centric navigation pipeline ``\textbf{VL-Explore}'' for simultaneous exploration and target discovery by UGVs in unknown environments. It requires no additional sensor modalities other than RGB feed for zero-shot active navigation.

\item We design a novel evaluation metric ``\textbf{Entropy Preserving Score}'' ($\EP$) to measure the exploration efficiency,
% mitigating the limitations of existing evaluation methods (\eg, SPL \JI{[?]}) that they often generate wildly different scores for the same navigation performance in different experimental environments. \JI{clarify}.
improving the consistency of score across different experimental environments hence providing better comparability across different systems.

\item We perform extensive experiments with the proposed system and compare its performance against other systems, demonstrating the effectiveness of the proposed system in real-world environments.

\end{enumerate}

\begin{figure*}[t]
    \centering
    \includegraphics[width=\linewidth]{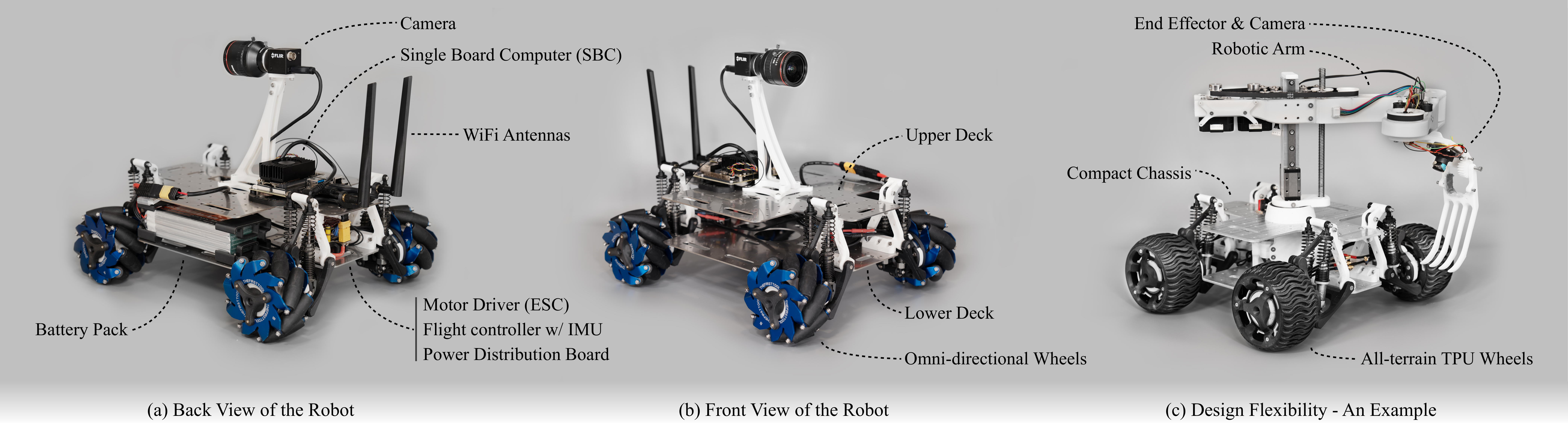}
    \caption{Our \ROVER ~ platform designed for \BRAND ~ is shown: (\textbf{a}) Back view, showing the single board computer (SBC) and electronics stack including a brushless motor speed controller, a flight controller with IMU, and a power distribution board;
    (\textbf{b}) Front view, showing the camera for zero-shot navigation; (\textbf{c}) An alternative design, demonstrating the flexibility of this platform, configured with revised wheels, chassis, and additional manipulators for potential field robotics applications.}
    \label{fig:system_design}
    \vspace{-2mm}
\end{figure*}

\section{Related Work: Zero-shot Learning and Vision-Language Navigation}\label{sec:background}

Zero-shot learning enables autonomous robots to identify unseen objects and make navigation decisions in unfamiliar environments~\citep{Guan2024LOCZSONLO}. Researchers use visual features~\citep{yang2018visual}, knowledge graphs~\citep{pal2021learning}, semantic embeddings~\citep{zhao2023zero}, and spatial appearance attributes~\citep{ma2024doze} to encode information of object classes into the search space. More recently, text-based descriptions or human instructions are combined with vision to improve servoing, as demonstrated in LM-Nav~\citep{shah2023lm} and InstructNav~\citep{long2024instructnav}. Among CLIP~\citep{radford2021learning}-based frameworks, CoW~\citep{gadre2022clip} presents a trajectory planning strategy using frontier-based exploration. ClipNav~\citep{dorbala2022clipnav} designs a \textit{costmap} to further improve obstacle avoidance during exploration. Other approaches such as ESC~\citep{zhou2023esc} and VLMaps~\citep{huang2023visual} use prompts to translate action commands into a sequence of open-vocabulary navigation tasks for planning. Besides, CorNav~\citep{liang2024cornav} includes environmental feedback in the zero-shot learning process to dynamically adjust navigation decisions on the fly.
% Besides, ESC~\citep{zhou2023esc} adopts a prompt-based grounding and an LLM for room and object reasoning and thus facilitates unknown scene understanding.

Contemporary works on vision-language navigation (VLN) focus on enabling robots to understand language instructions for interactive task planning. VLN algorithms integrate visual information (\eg, recognizing objects, obstacles) and language instructions to plan a trajectory or action sequence for task execution. This is achieved by first generating semantic tags into the SLAM pipeline~\citep{huang2023visual,Gadre2022CoWsOP, Guan2024LOCZSONLO} to generate a 3D map. Subsequently, the resulting augmented map is used to support online navigation to perform zero-shot semantic tasks~\citep{yokoyama2023vlfmvisionlanguagefrontiermaps}. These works primarily focus on adding semantic information to existing mapping and exploration techniques~\citep{huang2023visual,krantz2020navgraph}, but do not participate in the path planning process.
% Zero-shot learning does not participate in guiding the robot during the initial mapping run. The performance of task execution with pre-generated maps on a dynamically changing scene remains to be validated.

In contrast, \textit{node graph} based VLNs~\citep{anderson2018visionandlanguagenavigationinterpretingvisuallygrounded, kiran2022spatialrelationgraphgraph,Savarese-RSS-19} construct a graph of the active map where each node represents a free location and edges stand for directly navigable paths. Dynamic expansion of a node graph is then achieved with the help of precise localization, typically achieved by additional sensors. Experiments show that these algorithms suffer from miscorrelating existing nodes when revisiting an explored area, and also in dynamically changing scenes. To address this, researchers are exploring zero-shot learning aided observation of unknown spaces~\citep{yokoyama2023vlfmvisionlanguagefrontiermaps}. Techniques like odometry, stereo depth, and LiDAR sensors are used alongside a pretrained vision-language model to achieve simultaneous mapping and exploration -- demonstrating a clear advantage regarding the efficiency of reaching the prompted goal.

The existing zero-shot learning and VLN approaches rely on dedicated sensors to construct a global map on the first run. Subsequent motion decisions are generated by a traditional trajectory planner. Online planning and navigation in dynamic and unknown spaces without a prior map is still an open problem, which we attempt to address in this paper.

% \YZ{
%     Reviewers might expect us to mention autonomous driving as related work. We might need to clarify the differences to avoid their confusion.
% }

%\JI{Perhaps we can talk about the table here and mention where \BRAND ~ Fits}

% %\section{UGV Platform: Mechanical Design}
% \section{Rover Master: System Design}
% We develop a novel robotic platform ``\ROVER'' (Fig.\ref{fig:system_design}). This platform is designed to be a lower cost yet higher performance alternative to the \textit{TurtleBot} \TODO{Ref.}. It is completely designed from scratch, with most of its parts designed for 3D-printing.

% Both the \textit{TurtleBot} and our \ROVER ~ platforms are targeted to fast prototyping of research or engineering projects. Such robotic platforms have been widely used in various research fields and provided significant convenience towards the users.

%\section{Rover Master: System Design}
%Current UGV platforms such as the \textit{TurtleBots}~\citep{turtlebot} are widely used in various research fields and provide significant convenience to users. However, they often come with a high price tag while lacking features and performances demanded by modern AI applications.

%\draft{ Struggling to find a suitable platform with a reasonable price tag, we set off to develop a new platform that not only meets the needs of this project, but might also come handy for other researchers with similar requirements.}

% We designed and developed a novel UGV named ``{Rover Master}'' as low-cost yet higher performance alternative to the \textit{TurtleBot} \TODO{Ref.}.

\section{\ROVER: Platform Design}
We develop a novel robotic platform ``\ROVER''. The platform is designed to be flexible with drive types, chassis sizes, sensor and actuation add-ons, so it can adapt to the needs of prospective research. The major sensors and actuation components are shown in Fig.~\ref{fig:system_design}; it includes a monocular RGB camera and a 2D LiDAR for exteroceptive perception. Four independent wheel assemblies are responsible for actuation; each wheel assembly is modular and self-contained, \ie, it consists of a gearbox, a brushless DC motor, and a suspension system; see Fig.~\ref{fig:suspension:cad}b. It also includes a pseudo-odometer that uses telemetry data from an electronic speed controller (ESC), which drives the motors. Additionally, the driver stack consists of a flight-controller module originally designed for quadcopters that reports onboard sensory data to the host computer for planning and navigation. 

Unlike existing platforms like the \textit{TurtleBots}~\citep{turtlebot}, \ROVER is designed with an open and spacious chassis to facilitate easy adaptation to any single board computer (SBC). In our setup, we configured the platform to: (\textbf{i}) deliver sufficient computational power to handle a general-purpose vision-language model and other computationally intensive tasks; (\textbf{ii}) have enough mechanical stability to hold the camera in an elevated position without excess vibrations even on uneven surfaces; and (\textbf{iii}) support 3-DOF motions (forward/backward, sideways and rotation), whereas most existing UGV systems have only 2-DOF: forward surge and twist rotation.

\subsection{Task Specific Design for VL-Explore}
We customize the \ROVER ~ platform with a configuration that supports real-time VLN capability for 2D environments. An omnidirectional drive system is chosen to account for the moderate surface irregularities. The four wheels are connected to separate brushless DC motors via a planetary gearbox with a reduction ratio of $16$:$1$. This allows the robot to travel at higher speeds and easily overcome moderate obstacles. A throttle limit of $20$\% is imposed to ensure operational safety, capping the robot's maximum speed at approximately $2$\,meters per second. 

\begin{figure}[t]
    \centering
    \includegraphics[width=\columnwidth]{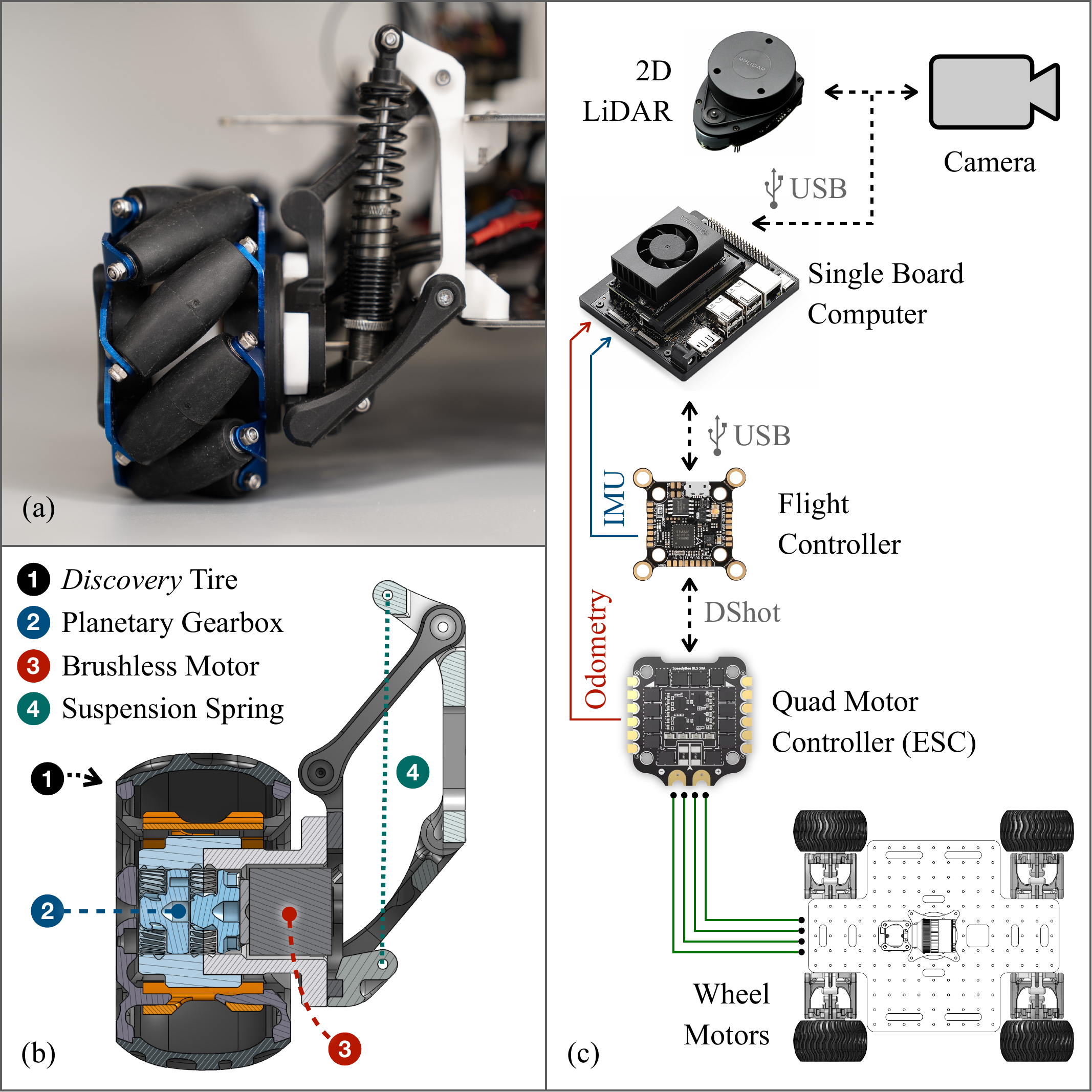}%
    \vspace{-1mm}
    \caption{
        The wheel hub and suspension design are shown on the left: (\textbf{a}) suspension system coupled with the omnidirectional wheel; and (\textbf{b}) cross-section view of the `discovery wheel', in-wheel drive system, and suspension links, assembled and rendered by CAD software. The connection diagram is shown on the right: (\textbf{c}) the dashed line represents physical connection, while solid lines represent logic functions.
    % \JI{make it more compact, remove the empty spaces} [DONE]
    }%
    \label{fig:suspension:cad}
    \label{fig:connection}
    \vspace{-1mm}
\end{figure}

As shown in Fig.~\ref{fig:suspension:cad}b, the gearbox and brushless DC motor are integrated into the wheel hub, optimizing the spatial efficiency and smoothness of the drive system. Moreover, the computing pipeline is powered by an Nvidia {\tt Jetson Orin} module. A {\tt FLIR} global shutter RGB camera, lifted approximately $30$\,cm above ground, is used for visual perception. This design ensures that the optical center of the lens aligns with the geometric center of the robot. Besides, the \textit{heading} value from the flight controller's IMU facilitates $360^{\circ}$ scanning capabilities. A 2D LiDAR is mounted at the front of the upper deck as is shown in Fig.~\ref{fig:system_design}b. The LiDAR is only used for safety and visualization purposes and does not influence VLN's decision-making. Further details regarding the LiDAR's utility are explained in \SUPPLEMNTARY{Sec.4}.

\subsection{Features and Capabilities}
One unique advantage of our \ROVER ~ system design is the compactness of its wheels and motor assembly. The four-wheel independent suspension extends its application to uneven surfaces; when using the \textit{Discovery Wheel} (TPU) shown in Fig.~\ref{fig:system_design}c and Fig.~\ref{fig:suspension:cad}b, it performs well on grasslands, rocky pavements, and even sand (with reduced maneuverability). The suspension system can also be configured to filter out surface bumps and provide better stability for onboard sensors.

\newcommand{\dRow}[1]{\multirow{2}{*}{#1}}
\newcommand{\acronym}[1]{\textbf{\small{[\,#1\,]}}}

\begin{table}
    \centering
    \renewcommand{\arraystretch}{1.1}
    \caption{Comparison of \textit{Rover Master} with other standard UGV platforms is shown; the acronyms: \acronym{DF} Differential, \acronym{OD} Omni-directional, \acronym{CFG} Configurable.}
    \vspace{-1mm}
    \resizebox{\columnwidth}{!}{
    \begin{tabular}{cccccc}
    \Xhline{2\arrayrulewidth}
    \dRow{Platform} &  Drive  & \dRow{SBC} & Onboard & \dRow{Add-ons} & Est. Cost  \\
                    &  Type   &            & Sensors &                & \footnotesize{(USD)} \\
    \Xhline{2\arrayrulewidth}
    TurtleBot3  & \dRow{\acronym{DF}} & \dRow{RPi4} &  PiCamera, IMU, & \dRow{Limited} & \dRow{$1500$} \\
    (Waffle Pi) & & & LDS laser & & \\ \hline TurtleBot4 & \dRow{\acronym{DF}} & \dRow{RPi4} & Stereo camera, & \dRow{Limited} & \dRow{$2100$} \\
    (Standard)  & & &  IMU, RPLiDAR & & \\\hline
    \dRow{\Large{$^\textit{~Rover}_\textit{Master}$}}
                & \acronym{DF} & \dRow{\acronym{CFG}} & RGB camera, & \dRow{\acronym{CFG}} & $\mathbf{\z650}$ \\\cline{2-2}\cline{6-6}
                & \acronym{OD} & & IMU, LiDAR & & $\z850$ \\ \Xhline{2\arrayrulewidth}
    %\multicolumn{6}{l}{\footnotesize{
    %    $^\dagger$ Configurable, can be equipped with various other SBCs such as Jetson Orin, Rock Pi, offering more computational power.
    %}} \\[-2mm]
    %\multicolumn{6}{l}{\footnotesize{
    %    $^\ddagger$ Including LiDAR modlue, monocular or stereo camera, robotic arm, actuated craws, \etc
    %}}
    % \multicolumn{6}{l}{\small{
    %     \textbf{\textit{Acronyms}}: \acronym{DF} Differential, \acronym{OD} Omni-directional, \acronym{CFG} Configurable.
    % }}
    \end{tabular}
    }
    \label{tab:cost}
    \vspace{-3mm}
\end{table}

The \ROVER ~ platform can be configured to switch between different \textit{drive types} (omnidirectional or differential), \textit{chassis sizes} (with parameterized CAD design), and task-specific \textit{actuation and sensory add-ons} (cameras, LiDARs, manipulators for both indoor and field robotics applications). The chassis plates are designed to house various additional sensors and actuators according to tasks.
% Additionally, it is configurable with multiple types of onboard computers, such as {Raspberry Pis}, Nvidia Jetson Modules, as well as mini X86 computers.
A comparison of \ROVER ~ with two widely used \textit{TurtleBot} UGV variants is presented in Table~\ref{tab:cost}.
\section{VL-Explore: Navigation Pipeline}\label{sec:arch}
Recent advancements in VLMs have demonstrated their potential for spatial reasoning and awareness capabilities~\citep{chen2024spatialvlm,dorbala2022clipnav}. Different from traditional approaches, this study explores the feasibility of incorporating VLMs directly into a robot's autonomy pipeline to facilitate real-time exploration and target identification. To this end, we introduce a modular pipeline that integrates stages for perception, planning, and navigation. The core computational components of the proposed VL-Explore ~ navigation pipeline are depicted in Fig.~\ref{fig:arch}.

The proposed pipeline comprises three main stages.
The \li{frontend} 
    processes raw input frames, divides them into tiles, and encodes these tiles into embeddings—numerical vectors representing semantic meanings; In our implimentation, the CLIP vision encoder is employed as the frontend.
Then, \li{middlewares}
    take the visual embeddings generated by the frontend as input and produce scores that carry specific semantic interpretations. These scores are typically derived by correlating the input embeddings with the middleware's internal database. The meanings of these scores can vary depending on the application's requirements. In this study, the middleware generates three types of scores: \textit{navigability}, \textit{familiarity}, and \textit{target confidence}.

Lastly, at the \li{backend},
    those scores from middlewares are used to make motion decisions. It is designed to be adaptable, allowing the integration of different algorithms tailored to various applications and environments. For this study, a minimal backend was implemented with three operational modes: basic navigation, look-around, and target lock. These modes are dynamically activated or deactivated based on the provided scores.

Notably, the proposed VL-Explore ~ pipeline is \textit{agnostic to} CLIP or any other VL backbone. OpenCLIP is used in our experiments given its open-source status and 3-4 years of maturity in various research fields, ensuring reproducibility. While a more recent multi-modal VLM could potentially enhance the performance of the suggested pipeline, the primary focus of this work is to validate the feasibility of integrating VLMs into a real-time navigation system. Thus, we chose to utilize a well-established model for better reproducibility.

\subsection{Visual Perception Frontend}
\begin{figure}
    \centering
    \vspace{-2mm}
    \includegraphics[width=\columnwidth]{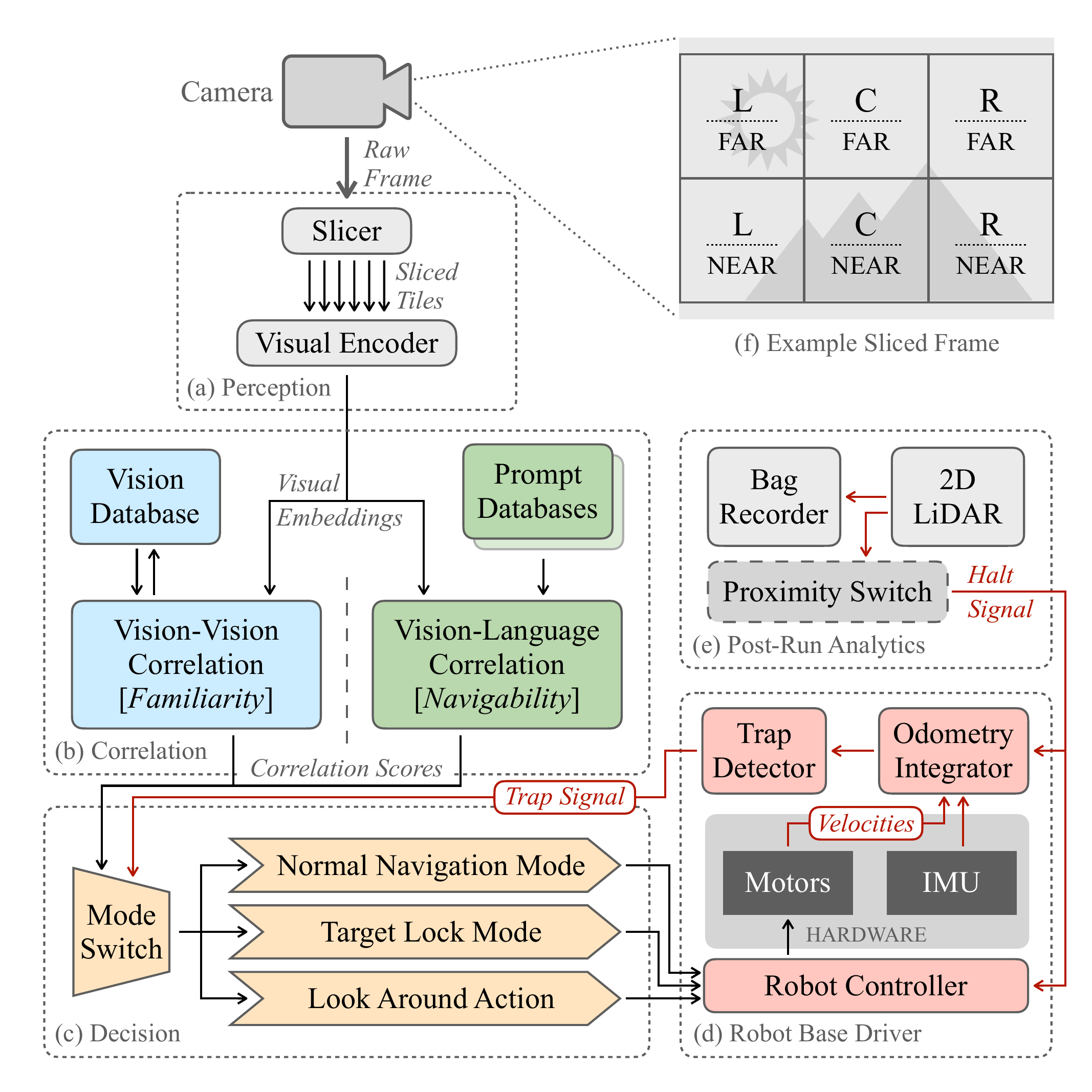}%
    \vspace{-3mm}
    \caption{
        An overview of the architecture of VL-Explore; the \textit{primary}
        data flow and \textit{supplementary} sensory feedback are marked by black and red arrows, respectively. The camera frame-slicing strategy is shown in (\textbf{f}); each frame is sliced into two rows (NEAR, FAR) and three columns (\underline{L}EFT, \underline{C}ENTER, and \underline{R}IGHT). Each of the six \textit{tiles} is encoded into a separate visual embedding vector for perception.
    }
    %\label{fig:segmentation-multi}
    %\label{fig:slicing}
    \label{fig:arch}
    \vspace{-3mm}
\end{figure}

In the frontend, raw camera frames are sliced into six tiles; the slicing strategies are discussed in \SUPPLEMNTARY{Sec.1}. Each tile represents a spatial location in the robot's FOV, which are scaled and processed by CLIP's visual encoder. As shown in Fig.~\ref{fig:arch}\,f, each frame is sliced into $N=6$ tiles and then rearranged into a tensor of shape $N\times 3 \times H \times W$, with $H$ and $W$ being the tile height and width in pixels. The encoder processes these inputs and generates an $N\times D$ embedding vector for each tile, where $D$ is the dimensionality of each prediction vector ($D=512$ in the CLIP model).

Additionally, the standard deviation for each tile is computed and combined with the model's predictions. This metric serves as an indicator of the \textit{amount of information} present in each tile, proving particularly useful in scenarios where the robot encounters feature-poor, uniformly colored objects such as walls, doors, or furniture. In such instances, an \textbf{abnormally} low standard deviation suggests that the vision encoder's output may lack reliability.

%\JI{you should refer to Fig5 and give some examples or explanations on how to interpret the embedding vectors}

\begin{figure*}
    \centering
    \includegraphics[width=\textwidth]{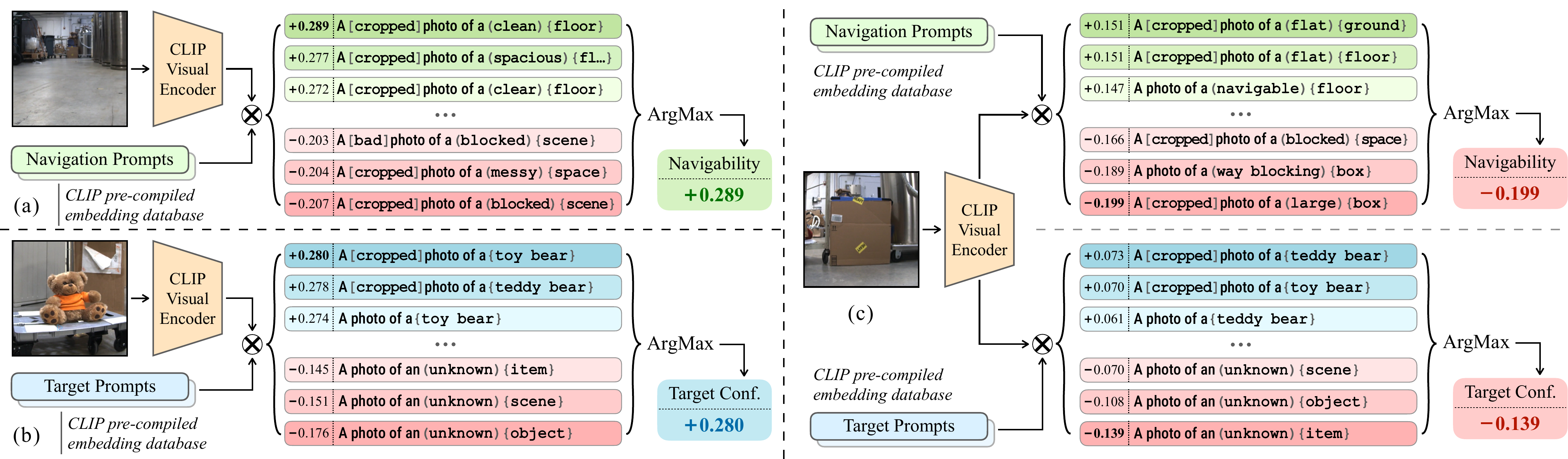}%
    \vspace{-2mm}
    \caption{Detailed examples of the proposed \textit{correlation middlewares} are illustrated; the circled cross symbol denotes the inner product of broadcasted vectors. (\textbf{a}) A navigable clean floor is encoded and correlated with the \textit{navigability} database, where green rows (positive prompts) yield higher scores than red rows (negative prompts); the resulting positive final score indicates the space is navigable. (\textbf{b}) A toy bear (the target) is encoded and correlated with the target database, where blue rows (positive prompts) produce higher scores than red rows (negative prompts); the resulting positive final score confirms the target's presence. (\textbf{c}) A paper box is encoded and correlated with both databases; it corresponds to neither a navigable space nor the target, both scores are negative, indicating the space is not navigable and no target is present. [Best viewed digitally at $2\times$ zoom.]}
    \label{fig:correlation}
    \vspace{-2mm}
\end{figure*}

\subsection{Navigability Middleware: Vision-Language Correlation}
To distinguish navigable spaces from non-navigable ones, we designed a set of \textit{positive prompts} describing clean and navigable environments, such as: \Prompt{A photo of a \SttPrompt{flat\DIV open\DIV wide\DIV clear} \ObjPrompt{floor\DIV ground\DIV hallway}}, and a set of \textit{negative prompts} describing spaces that are cluttered by obstacles, such as: \Prompt{A \DscPrompt{cropped\DIV bad\DIV imcomplete} photo of a \SttPrompt{blocked\DIV messy\DIV cluttered} \ObjPrompt{scene\DIV space}} and \Prompt{A photo of a \SttPrompt{large\DIV way blocking} \ObjPrompt{object\DIV item}}. As shown in Fig.~\ref{fig:correlation}\,a, a clear floor is identified as navigable space, while the cluttered scene in Fig.~\ref{fig:correlation}c is accurately categorized as non-navigable.

For target discovery, a similar set of text prompts is used to define the \textit{target} of a task. In our experiments, we use a toy bear (see Fig.~\ref{fig:env}\,b) as the discovery target due to its uniqueness in the scene. We design a set of prompts that describe the target, \eg~\Prompt{A photo of a \SttPrompt{brown\DIV toy} \ObjPrompt{bear\DIV teddy bear}}. We also design a set of negative prompts that describe generic objects to filter out false positives, \eg~ \Prompt{A photo of an \SttPrompt{unknown} \ObjPrompt{item\DIV scene\DIV object}}. As shown in Fig.~\ref{fig:correlation}\,b, the \textit{target prompt} accurately identified the toy bear, while negative prompts effectively suppressed false positives on unrelated objects, such as the paper box in Fig.~\ref{fig:correlation}\,c.

The symbols used in examples above belong to a custom designed prompt system. The same set of notations are used throughout this paper. The symbols are explained in detail in \SUPPLEMNTARY{Sec.2}.

The resulting scores for both navigability and target confidence were computed on a per-tile basis. As depicted in Fig.~\ref{fig:correlation}\,a-c, the CLIP visual encoder generated embeddings for each tile, which were then compared against a prompt database using inner products. The prompt database consisted of pre-encoded text prompts generated by the CLIP text encoder, organized into positive and negative categories. The resulting scores, ranging from $-1.0$ to $1.0$, were determined by the highest absolute score among the prompt matches.

By employing both positive and negative prompts in each database and selecting the final result based on their contrast, the correlation process becomes more robust against fluctuations in absolute correlation values. This approach is particularly beneficial in environments with varying lighting conditions or complex scenes, where all scores may drift upward or downward depending on the quality of the visual input. Moreover, this method eliminates the need for manually setting a fixed threshold, further enhancing its adaptability.

% \JI{
% reviewers would want to know how the {\tt Nav} correlation scores are calculated, some insights are missing here. I did not understand how the vision-language correlation happens}
% % \TODO{\YZ{Will add an algorithm block to explain this.}}
% \YZ{Added Fig.~\ref{fig:correlation}, please have a look!}
% \JI{Also, are we talking about Target, for target discovery cases only? How about the pure exploration case}

\subsection{Familiarity Middleware: Vision-Vision Correlation}\label{sec:arch/decision/familiarity}
In addition to navigability scores, a \textit{familiarity database} is accumulated in real-time to track previously explored spaces. It is constructed with visual embeddings ($512$ dimensional vectors) representing known spaces without storing or using actual images. An incoming visual embedding is considered ``known'' if its correlation score with an existing vector exceeds a predefined threshold. Each new embedding vector is incrementally merged into the familiarity database; we implement the following two strategies for this:
\vspace{1mm}
\begin{enumerate}[label={$\arabic*$.},nolistsep,leftmargin=*]
\item Averaging among all vectors that belong to a known point, this involves keeping track of the count of vectors already merged into a known spot ($s$):
$$
v_\text{next} = \frac{s}{s + 1} \cdot v_\text{prev} + \frac{1}{s + 1} \cdot v_\text{new}
$$

\item Performing a rolling average operation upon merging a new vector; this method does not need to keep track of the total count of already merged vectors. Therefore, the vector tends to lean towards newly inserted vectors and gradually ``forget'' older ones. The ``rate of forgetting'' can be controlled by a factor $\lambda$ (a.k.a decay factor):
$$
v_\text{next} = (1 - \lambda) \cdot v_\text{prev} + \lambda \cdot v_\text{new}
$$
\end{enumerate}
When no \textit{known vector} exists in the database, the incoming vector is inserted as a new data point. Eventually, a familiarity score is generated for each perception vector, guiding navigation by encouraging the robot to prioritize unexplored areas over revisiting familiar ones.

\begin{figure}[t]
% no need for additional padding
    \centering
    \includegraphics[width=0.8\columnwidth]{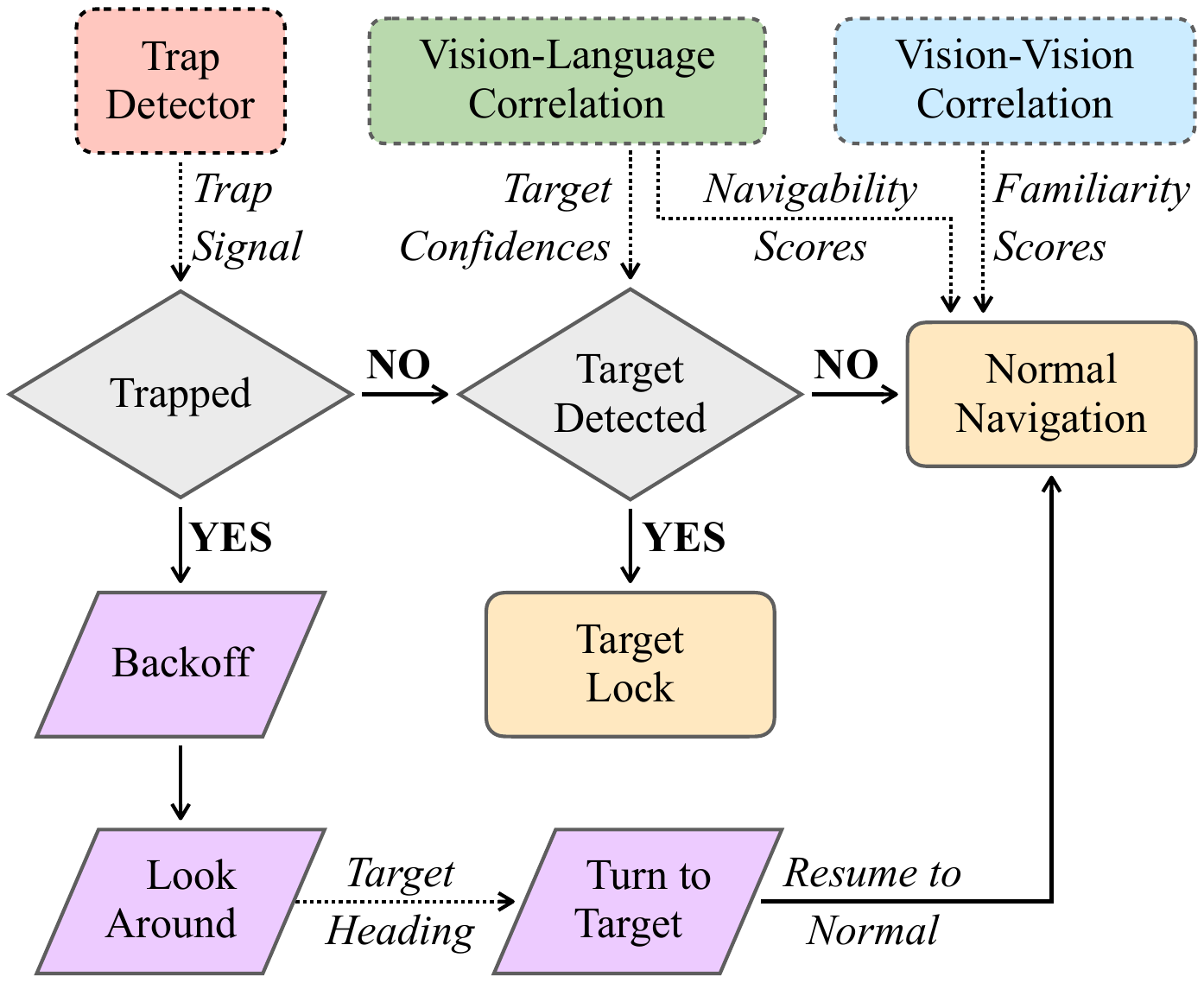}%
    \vspace{1mm}
    \caption{
        A state diagram illustrating the decision-making process, providing a detailed view of Fig.~\ref{fig:arch}c; blocks with dashed borders are other computational blocks in Fig.~\ref{fig:arch}. Dashed arrows represent data/signal propagation, while solid arrows denote conditions and transitions between states.
    }%
    \label{fig:state-diagram}
    \vspace{-3mm}
\end{figure}

\subsection{Navigation Decision Backend}\label{sec:arch/decision}
Lastly, the decision module generates motion commands according to the information provided by the perception and correlation systems. Specifically, a ``motion mixer'' is introduced as the baseline correlation-to-motion translator. It takes all aforementioned scores for each tile (\ie, \textit{navigability},
\textit{familiarity}, and \textit{standard deviation}) into consideration and makes an intelligent decision. The motion mixer generally prioritizes highly navigable yet less familiar locations while avoiding areas with minimal texture, indicated by low standard deviation values. To handle non-trivial scenarios, the decision module incorporates two additional functionalities: (1) \textit{trap detection}, enabling the robot to identify and escape from potential dead-ends, and (2) \textit{look-around}, allowing it to reorient itself in complex environments. The overall decision-making process and state transitions are illustrated in Fig.~\ref{fig:state-diagram}.

% \JI{
%     Need to end this section with a state machine diagram for the navigation decisions. It seems very high-level.
% }
% \YZ{
%     The state machine is actually a ``multiplexer'' shown in Fig.~\ref{fig:arch}c. Would you like to have it in a separate figure?
% }
% [DONE]
\vspace{1mm}
\noindent
\textbf{Trap Detection.}
In certain scenarios, the robot may encounter a ``dead end'', where no navigable path is visible within its FOV. In rare instances, the vision-language model may generate false positive \texttt{Nav scores} for scenes it does not adequately comprehend. For example, the lack of salient features can lead to incorrect positive scores when the camera is positioned too close to a plain white wall. Such situations are defined as ``trapped'' states, which can occur under two conditions: (a) the proximity switch asserts a halt signal for a specified duration, or (b) the cumulative travel distance, as measured by odometry, falls below a defined threshold over a given period. Empirically, the system flags the robot as trapped if it travels less than $0.2$ meters within the past $5$ seconds.

\begin{figure}
    \centering
    \includegraphics[width=\columnwidth]{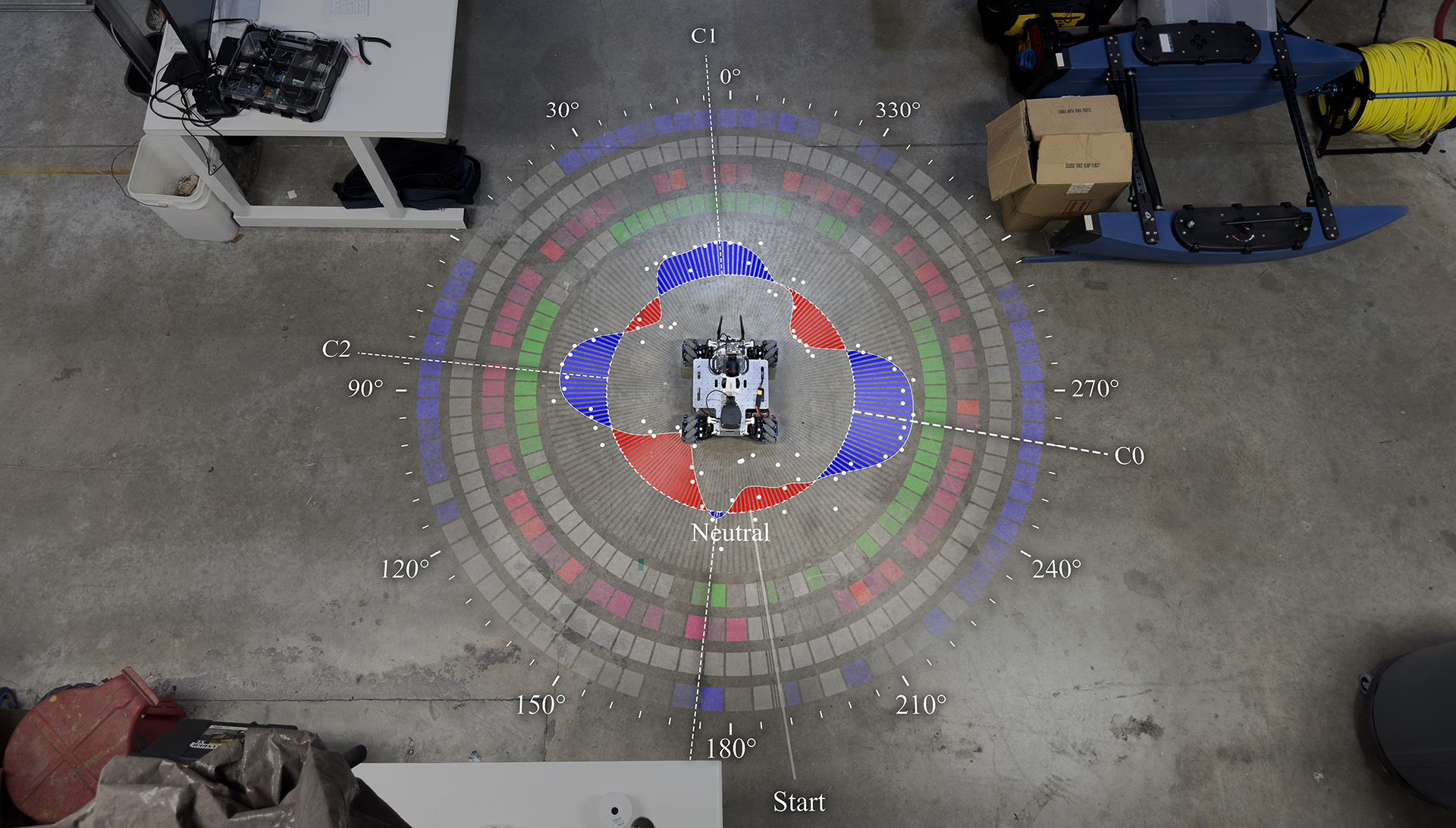}%
     \vspace{-2mm}
    \caption{Rendering of a \textit{look around} operation overlaid on a birds-eye view of the robot; it was able to identify navigable paths (blue bars) apart from obstacles (red bars) based on the proposed visual perception pipeline. The candidate headings are annotated as $C_i$. After the look-around operation, the robot selected $C_0$ as its next heading according to the area of free space. Details on the color codes and symbols are in Fig.~\ref{fig:exp}.
    }
    \vspace{-3mm}
    \label{fig:visual-homo}
\end{figure}

\vspace{1mm}
\noindent
\textbf{Look Around.}\label{sec:arch/look-around}
Due to the limited FOV from a single camera, the robot could only see objects in front of it, limiting the amount of information available to the robot for efficient navigation and exploration. To address this, a ``look around'' mechanism (see Fig.~\ref{fig:visual-homo}) is introduced to enable the robot to gain situational awareness by performing a $360^\circ$ rotation while collecting \texttt{Nav scores} associated with different headings. A Gaussian convolution is then applied to these scores to identify the most navigable direction. When recovering from a ``trapped'' state, the \textit{look around} mechanism prioritizes a direction different from the original heading by a linear factor $k$, rewarding headings that deviate from the initial orientation. Additionally, a \textit{look around} behavior is triggered at the start of a new mission to ensure the robot identifies the most promising path for exploration.

\section{Experimental Results and Analyses}\label{sec:exp}

\subsection{Experimental Setup}
Real-world experiments were conducted in two different environments as depicted in Fig.~\ref{fig:env}\,a. The environments were selected for their cluttered layout and complex details, which include numerous obstacles, potential traps, and loops, providing a challenging setting for comprehensive analysis. As shown in Fig.~\ref{fig:env}, the robot was tasked with exploring the space while searching for a designated target -- a toy bear approximately $20$\,cm tall and $10$\,cm wide, chosen for its distinctive appearance within the test scene. For each task, the source (robot's starting position) and destination (target location) were selected from predefined locations as labeled in Fig.~\ref{fig:env}\,(b) and (d).

%As shown in Fig.~\ref{fig:env}, the robot is tasked to explore the lab space while looking for a specific target, which is a toy bear (approximately $20$\,cm tall and $10$\,cm wide); it is chosen to ensure its uniqueness in the test scene. In each task, the source (robot's starting position) and destination (target location) are chosen from the five regions (\eg, SW, C, NW, NE, SE) labeled in Fig.~\ref{fig:env}c.

\begin{figure*}
    \centering
    \includegraphics[width=\textwidth]{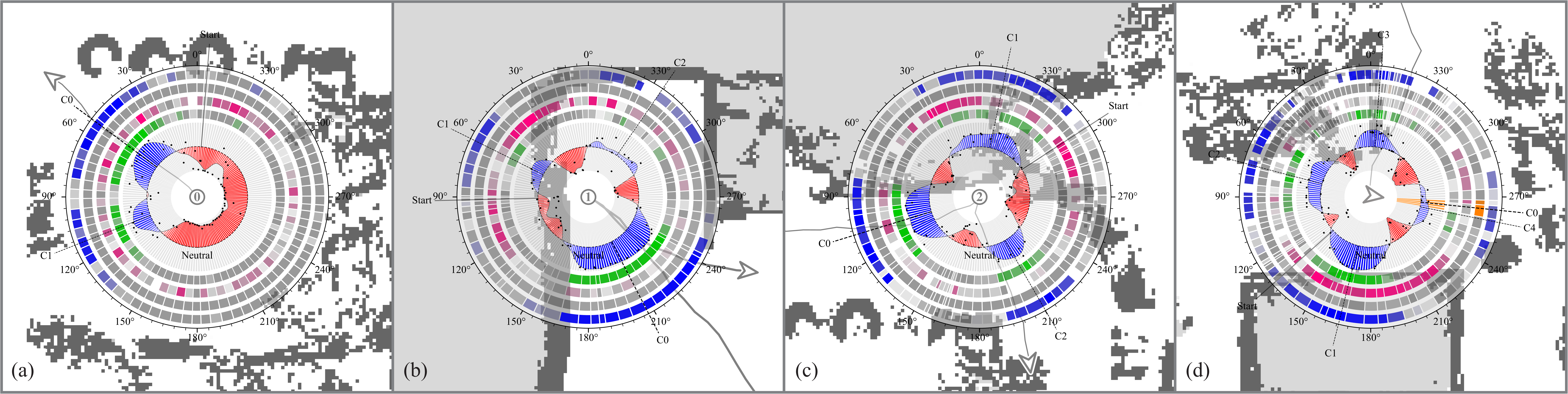}
    \includegraphics[width=\textwidth]{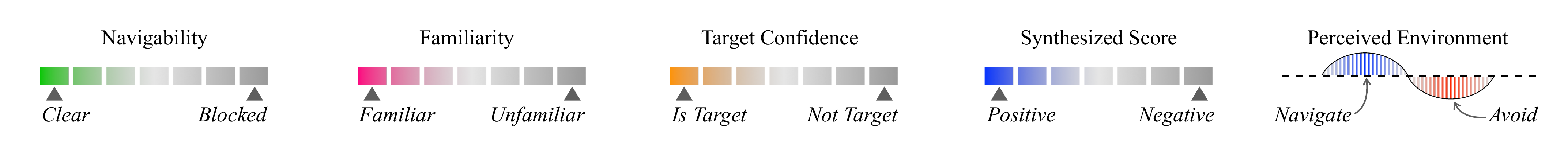}%
    \vspace{-3mm}
    \caption{Visualizations of look-around operations during a demonstrative task; candidate directions are depicted as dashed lines ($C_i$), where candidates with smaller indices are assigned a higher priority. The plots are generated from correlation scores without utilizing additional sensory data. (\textbf{a}) Initial look-around at the center of the map: the system does not incorporate familiarity scores since the familiarity database is uninitialized; thus, the navigability score predominantly influences the robot's decision. (\textbf{b}) The robot encountered a dead-end and was temporarily trapped; a look-around operation enabled it to identify a navigable path and resume exploration. (\textbf{c}) The robot was trapped due to a false positive navigability perception caused by a transparent object; with a look-around, the system successfully recovered from the false positive and continued its exploration. (\textbf{d}) The target was located nearby in this scenario; through the look-around, the robot was able to identify the target and assign it as the highest-priority candidate. [Best viewed digitally at $2\times$ zoom for clarity.]
    }
    \label{fig:exp-look-around}
    \vspace{-3mm}
\end{figure*}

\begin{figure}
  \centering
  \includegraphics[width=\columnwidth]{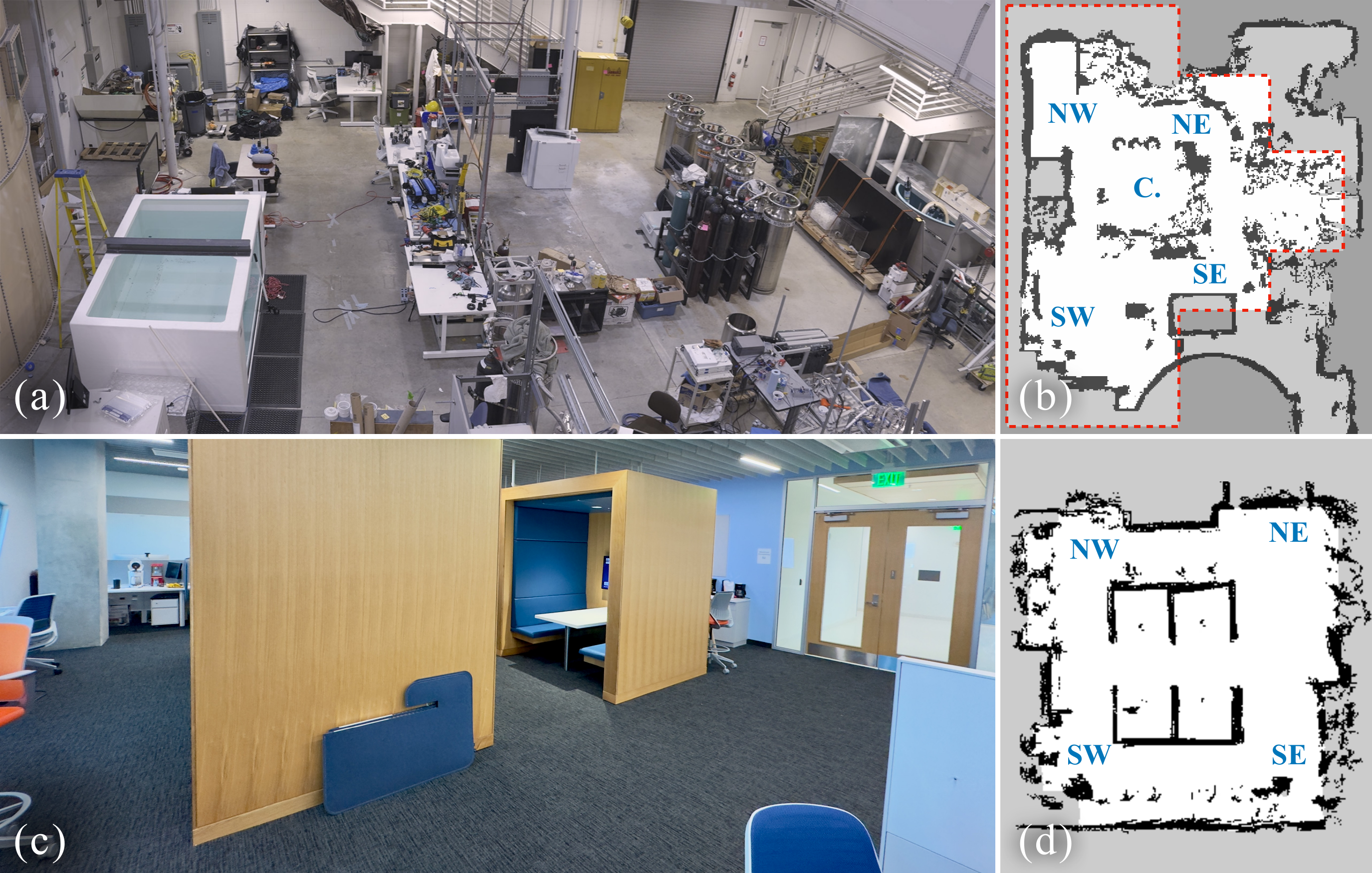}%
  \vspace{-2mm}
  \caption{The environment used for real world exploration and discovery experiments: (\textbf{a}) Overview of the machine hall taken from the east side. (\textbf{b}) 2D map of the experimental space, with task locations annotated in blue; the area utilized for the experiments is enclosed within the dashed red line. (\textbf{c}) An office scene taken from the \SE corner; (\textbf{d}) 2D map of the office scene, with task locations annotated in blue.
  % [DONE]
  % \JI{for (a), better take a top-view photo from the second floor to show the space, that resembles the map; the current photo is not informative.}
  }
  \label{fig:env}
  \vspace{-2mm}
\end{figure}

\subsection{Performance Evaluation}
\noindent 
\textbf{Evaluation criteria.} Our experimental trials are designed to evaluate \textit{efficiency} of exploring an area and \textit{success rates} of finding a target with no prior map or knowledge about the target. To quantify these, we use the total distance traveled (trajectory length) before approaching the target as the core metric. Instead of the total duration of the task, the travel distance metric is agnostic to the CPU/GPU performance of the onboard computer and the capability of the mechanical driving system, which are considered external factors that can be improved independently.

\vspace{1mm}
\noindent 
\textbf{Algorithms for comparison.} We compare the performance of the proposed VL-Explore system with six widely used map-traversal and path-finding algorithms. For fair comparisons, a novel 2D simulation framework, \textbf{RoboSim2D}, is developed. It utilizes 2D LiDAR maps from recorded scans of the same environment used in real-world experiments. Additionally, the size of the simulated robot was configured to match the dimensions of the physical robot, ensuring consistency across simulations and the real platform. See the \underline{supplementary materials} (Sec.~5) for more details on the algorithmic implementation and simulation results.

%\subsection{Criteria of Failure}
\vspace{1mm}
\noindent
\textbf{Criteria of failure.} Due to practical limitations, a mission is considered a failure if the distance traveled exceeds a predefined upper limit. For real-world experiments conducted in the test area (approximately $12 \times 16$ m), this limit is set to $100$ meters. In simulations, the limit is set to $1,000$ meters for random walk, wall-bouncing, and wave-front algorithms. For Bug algorithms, loop detection is employed to identify endless loops, which are classified as failures. Since failed missions yield infinite travel distances, they are excluded from the metrics presented in Table~\ref{tab:overall}, they are instead reflected in the overall success rate of an algorithm.

\vspace{1mm}
\noindent
\textbf{Evaluation metrics.} As observed in our simulation results, for a given amount of information, the success rate $R$ is inversely proportional to the normalized path length $L$. This relationship is expressed by a commonly adopted performance metric for VLN systems, known as \textit{Success weighted by Path Length} (SPL)~\citep{anderson2018evaluation}, defined as:
\begin{equation}
\text{SPL} = \frac{1}{N} \sum_{i=1}^N S_i \frac{l_i}{\max(p_i, l_i)}.
\label{eq:SPL}
\end{equation}

\noindent
Note that, for each source-target combination, the first contact distance of WaveFront simulation is used as the baseline distance $D_\text{baseline}$. With the choice of $l = D_\text{baseline}$, we have $p_i > l, \forall p_i \in \{p\}$; hence, the above equation is simplified to:

\begin{equation}
\text{SPL}
  = \frac{1}{N} \sum_{i}^{S_i = 1} \frac{D_\text{baseline}}{p_i}
  = \frac{N_s}{N} \sum_{i}^{S_i = 1} \frac{L_i}{N_s}
  = \,R\, \cdot \,\overline{L}
\end{equation}
\begin{equation}
\text{s.t.}
\left\{\scalebox{.9}{$
    \setlength\arraycolsep{0pt}
    \begin{matrix*}[l]
        \,\,L_i & = \dfrac{D_\text{baseline}}{p_i}
        & ~ - ~ \text{Inverse Relative Distance} \\[8pt]
        \,\,N_s & = \sum_{i}^{S_i = 1} 1
        & ~ - ~ \text{Number of Success Runs} \\[4pt]
        R & = \dfrac{N_s}{N}
        & ~ - ~ \text{Success Rate} \\[4pt]
        ~\, \overline{L} & = \sum_{i}^{S_i = 1} \dfrac{L_i}{N_s}
        & ~ - ~ \text{Mean Inverse Path Length} \\[4pt]
    \end{matrix*}
$}\right.
\label{eq:definitions}
\end{equation}

\noindent
As a consequence, the SPL metric equates to an inverse proportional relation between two quantities, namely the success rate $R$ and the mean inverse path length $\overline{L}$, as:

\begin{equation}
\text{SPL} = \overline{L} \cdot {R}.
\end{equation}

\begin{figure*}[t]
    \centering
    \includegraphics[width=\textwidth]{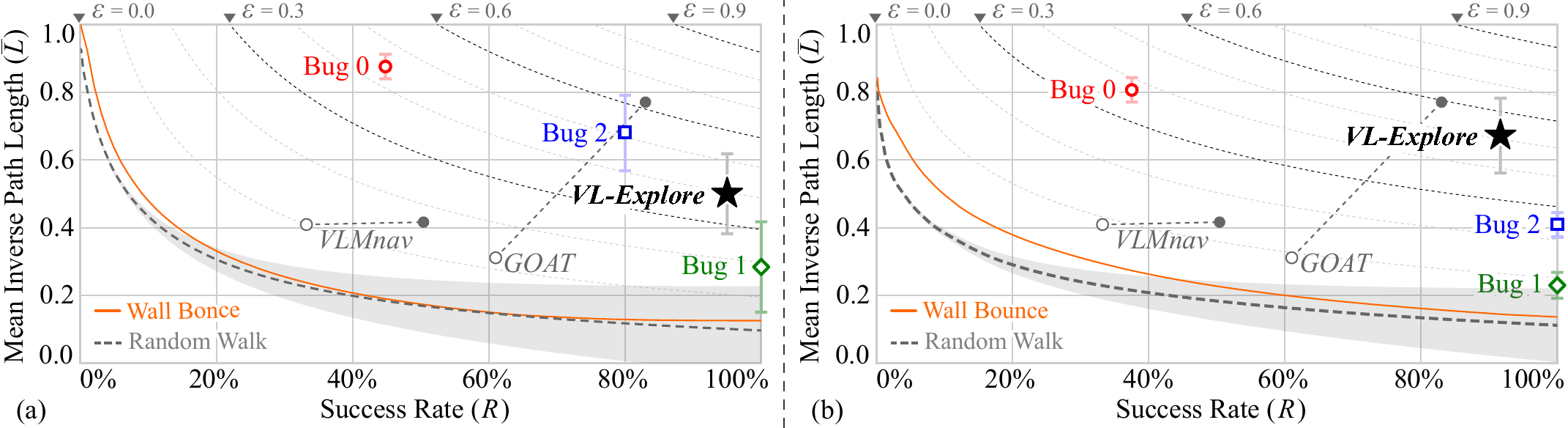}
    \caption{Performance of VL-Explore compared to map-traversal methods (\ie~ random walk, wall bounce)~\citep{pang2019randoomwalk}, path-finding (Bug) algorithms~\citep{lumelsky1986dynamic,lumelsky1987path} and SOTA VLN methods (\ie~ VLMnav and GOAT)~\citep{goetting2024endtoend,chang2023goatthing} for UGV navigation. {SOTA VLN methods are represented by a solid gray dot (all systems enabled) and a gray circle (certain systems disabled to match with information required by VL-Explore). Detailed explanations are provided in Table.\ref{tab:overall}}. (a) Large industrial scene; (b) Mid-size office scene. The equipotential lines labeled as $\EP = x$ are defined by Eq.~\ref{eq:exp/epsilon}. % Note that the \textit{relative efficiency} is calculated as an inverse metric to the normalized trajectory path length from baseline; details are provided later in Table.~\ref{tab:results}.
    % \JI{explain the relative efficienty}
    %VL-Explore consistently outperforms map-traversal algorithms such as wavefront and random walk. It also achieves comparable and often better performance than the Bug algorithm~\citep{lumelsky1986dynamic,lumelsky1987path}-based path planners; note that they have an inherent advantage of relying on prior knowledge of the target location, with Bug1 further requiring precise localization.
    }
    \label{fig:overall}
    % \vspace{-5mm}
\end{figure*}

\noindent
However, experimental data reveal that the curvature of the SPL function does not align well with the observed $R$--$\overline{L}$ curve, shown later in the experimental results (Fig.~\ref{fig:overall}).
That is, for the same exploration algorithm, setting a different failure criterion will result in a different SPL score. To better model the efficiency of an algorithm from observed data, {\textbf{we propose a new metric ``\textit{Entropy Preserving Score}'' ($\EP$)}} to represent the amount of information available to a given algorithm. It has a range of $[0, 1]$. When $\EP=0$, the algorithm operates without environmental information and lacks sensing capabilities beyond collision detection, as seen in methods like random walk and wall bounce. As $\EP$ approaches $1$, it gains more information or has higher environment sensing capability, and the performance is expected to increase accordingly. Considering the impact of $\EP$, the revised equation expands to:

\begin{equation}
f_1(\EP) = f_2(\overline{L}) \cdot f_3(R)
\label{eq:exp/epsilon}
\end{equation} \vspace{-3mm}
\begin{equation*}
\text{s.t.} ~ f_n(x) = k_n \cdot x ^ {\,p_n} + t_n
\end{equation*}

\noindent
where $H_n = [ k_n ~~ t_n ~~ p_n ]$ are hyper-parameters specific to an environment. \ie~A specific environment can be characterized using the collection of 9 hyperparameters:

\begin{equation}
H
= \left[
  \begin{matrix}
    H_1 \\
    H_2 \\
    H_3
  \end{matrix}
\right]
= \left[
  \begin{matrix}
    k_1 & t_1 & p_1 \\
    k_2 & t_2 & p_2 \\
    k_3 & t_3 & p_3
  \end{matrix}
\right].
\end{equation}

These hyper-parameters are fitted to the $\overline{L} - {R}$ curve derived from random-walk simulation results in the given map, plus an additional boundary condition $f_1(1.0) = f_2(1.0) \cdot f_3(1.0)$, which is the ideal case when the algorithm has all information available and achieves a perfect success rate with optimal travel distance.

\begin{table*}
  \caption{
    Quantitative performance of VL-Explore and other algorithms in comparison. Metrics used are defined in Eq.~\ref{eq:definitions}. \\
    \phantom{\textbf{Table 4.~}} The acronyms: S.R. - Success Rate $R$;\quad P.L. - Mean Inverse Path Length $\overline{L}$.
    %Path lengths are normalized per row, as detailed in Table~\ref{tab:results}, using the first contact distance of the Wave Front simulation as the baseline. \JI{dont understand the caption}%Algorithms are divided into two sections by a double rule, with comparisons made within each section.
  }
  \includegraphics[width=\textwidth]{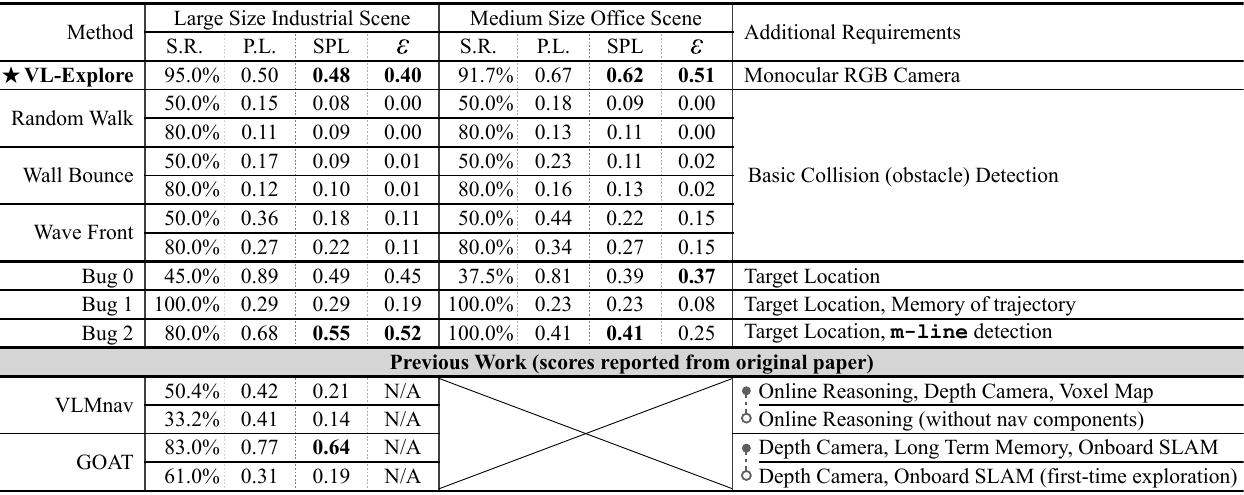}
  \label{tab:overall}
  \vspace{-3mm}
\end{table*}

\begin{figure*}
  \centering
  \includegraphics[width=\textwidth]{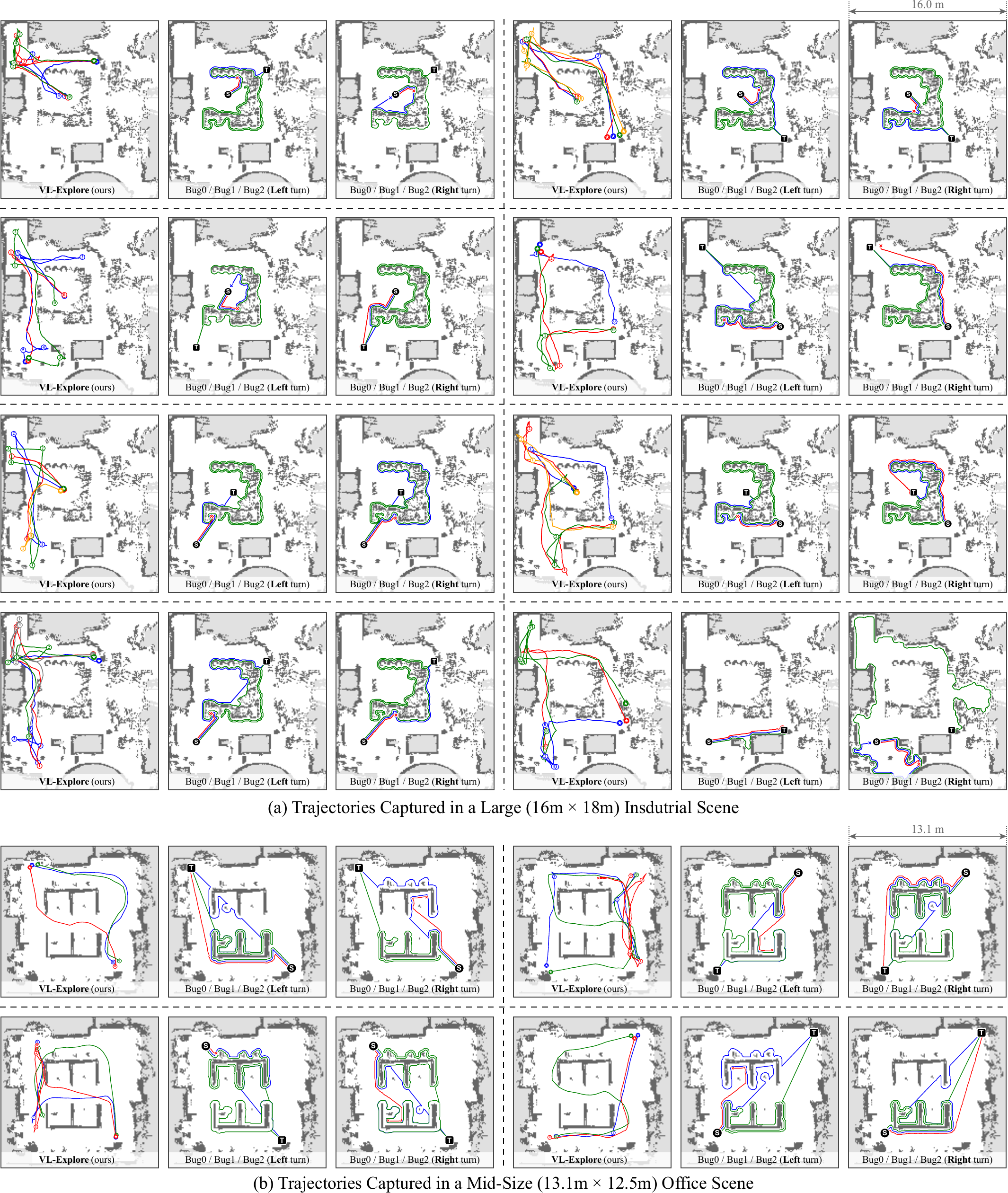}%
  \vspace{-1mm}
  \caption{
      Sample results of autonomous exploration and target discovery by VL-Explore compared to Bug algorithms. %\draft{Showing $\mathbf{8}$ out of $\mathbf{20}$ combinations. A complete matrix of all experiments is attached in the Appendix.} 
      The trajectories of each algorithm are overlaid on a 2D map of the test environment (north up). Circled numbers indicate the \textit{look around} operations of VL-Explore and the corresponding waypoints sequence it followed. For Bug algorithms, the red, green, and blue trajectories represent paths traversed by \textit{Bug0}, \textit{Bug1}, and \textit{Bug2}, respectively. Best viewed digitally at $2\times$ zoom for clarity; more results and analyses are included in \SUPPLEMNTARY{Sec.6}.
  }
  \vspace{-2mm}
  \label{fig:exp}
\end{figure*}

\subsection{Qualitative and Quantitative Analyses}

% Moved from the intro section
% We conducted extensive experiments to evaluate the proposed system against both map-traversal and path-finding algorithms. A performance comparison based on efficiency and success rates is shown in Fig.~\ref{fig:overall}, identifying equipotential lines labeled as $\EP = x$. In our evaluation, we observe that algorithms that use a similar amount of information tend to yield similar $\EP$ scores. A higher $\EP$ scores represents more information (\eg~ prior map of the environment, target location, additional sensory capabilities) available to the algorithm. With zero prior information of the environment and minimum sensing capabilities ($\EP = 0$), VL-Explore offers performance margins comparable and often better than Bug2. Note that Bug2 needs prior information on the target position and a precise localization to keep track of its {\tt m-line}. Overall, VL-Explore delivers  \textbf{$\mathbf{33\%}$ more efficient} source-to-target paths compared to Bug2 and achieves a \textbf{$\mathbf{10\%}$ higher success} than the Bug1 algorithm. Importantly, none of the evaluated planner and traversal methods have a semantic understanding of the scene, limiting their ability to adapt to dynamic environments. \JI{I thought we were moving this to the later sections, why is this discussion still here?}

We conduct extensive real-world experiments with over $60$ trials based on various combinations of source-target locations shown in Fig.~\ref{fig:env}. For the proposed VL-Explore system, three trials were performed for each origin-target pair. In comparison, $200$ trials were recorded for each random walk experiment, while $180$ evenly distributed headings were recorded for each combination of the wall bounce runs.

Samples of the trajectories traversed by VL-Explore and other algorithms are presented in Fig.~\ref{fig:exp}, overlaid on a 2D map of the environment. Under \textit{normal navigation mode}, VL-Explore exhibits a human-like motion strategy by navigating near the center of open spaces, enhancing exploration efficiency and reducing the likelihood of becoming trapped by obstacles. In contrast, path-finding algorithms often follow the contours of obstacles due to their lack of semantic scene understanding. During \textit{look-around operations} (see Fig.~\ref{fig:visual-homo} and Fig.~\ref{fig:exp-look-around}), VL-Explore accurately discriminates between navigable spaces and obstacles. Additionally, it demonstrates a preference for unexplored (unfamiliar) areas over previously visited (familiar) regions, contributing to improved efficiency compared to traditional map-traversal algorithms.

%Samples of trajectories traversed by the proposed VL-Explore system and other algorithms in comparison are shown in Fig.~\ref{fig:exp}, overlaid on a 2D map of the environment. In \li{normal navigation mo, VL-Explore is programmed to exhibit a human-like motion preference to navigate at the center of free spaces. This helps to improve the efficiency of exploration and reduce the chance of being trapped by obstacles. In contrast, the path-finding algorithms often travel along the contours of obstacles due to their lack of high-level understanding of the scene. In \li{look-around operations} (see Fig.~\ref{fig:visual-homo} and Fig.~\ref{fig:exp-look-around}), VL-Explore accurately discriminates navigable spaces apart from obstacles. Besides, it exhibits a general preference towards unexplored (unfamiliar) areas over previously explored (familiar) areas, contributing to a higher efficiency over the map traversal algorithms.

Out of $60$ trials in the industrial scene, 3 failure cases were recorded, yielding a $95\%$ overall success rate for VL-Explore. For the office scene, only 1 failure case was recorded out of $12$ trials, yielding a $91.7\%$ success rate. Among the failure cases, two were caused by the trajectory length exceeding the limit, while the other two were caused by the robot getting jammed by an obstacle, which was not detected by either the VLN pipeline or the LiDAR proximity switch.

The quantitative results categorized by source and target locations are presented in \SUPPLEMNTARY{Table.1}. To ensure comparability across different source-target pairs, the travel distance for each row is normalized by the baseline travel distance and then aggregated into Table~\ref{tab:overall}. For randomized algorithms, the success rate $R$ is a configurable hyperparameter. Sample results for $R = 50\%$ and $R = 80\%$ are provided to represent their underlying performance in Table~\ref{tab:overall}, with the corresponding equipotential curves previously shown in Fig.~\ref{fig:overall}. In contrast, deterministic algorithms (such as Bug variants) have a fixed success rate. Their performances are reported directly in Table~\ref{tab:overall} and visualized in Fig.~\ref{fig:overall} for comparison.

% As these results demonstrate, VL-Explore averages $2.66 \times D_\text{baseline}$ travel distance before reaching the target, while the next best score is from Wall Bounce, with $9.58 \times D_\text{baseline}$ at a significantly lower success rate ($50\%$). Note that, it achieves this performance gains despite having the same prior information as the map traversal algorithms. On the other hand, although the Bug* algorithms uses additional information about the environment map and target location, VL-Explore ($95\%$) still outperforms outperforms Bug1 ($6.34 \times D_\text{baseline}$) by trajectory lengths, and offers significantly higher success rates than Bug0 ($45\%$) and Bug2 ($80\%$) algorithms.

As these results demonstrate, with the same amount of information, the proposed system outperforms traditional map traversal algorithms by significant margins in success rates, trajectory lengths, SPL and $\EP$ scores. Despite the inherent disadvantage of operating without prior target knowledge or precise localization, the system achieves performance comparable to path-finding algorithms. Notably, in many scenarios, it surpasses path-finding algorithms in either trajectory efficiency or success rates. These results highlight the effectiveness of the proposed vision-language based pipeline for simultaneous exploration and target discovery tasks.

\subsection{Comparative Analyses of SOTA VLNs}
We also compare VL-Explore with SOTA VLN systems based on their features and prerequisites.
It is important to note that, due to different experimental setup and reporting criteria, a perfect quantitative comparison is not always feasible. For example, SEEK \citep{Ginting2024Seek} reported $0.65${\small$\pm0.35$} SPL for random walk, while our random walk simulation yielded $0.08${\small$\pm0.11$} SPL, showing a magnitude of difference. Preliminary analysis suggests that this discrepancy stems from SEEK's reliance on a node-graph map structure, which reduces the problem space to a finite set of graph nodes and edges. In contrast, VL-Explore operates continuously in an unknown environment, resulting in a substantially larger problem space, facilitating active robot navigation with no prior information or pre-compiled waypoints. This fundamental difference motivated us to introduce the $\EP$ metric, which helps reduce the discrepancies caused by varying prerequisites and experimental setups.

Among the vast volume of prior attempts in the field of vision-language based navigation or exploration, we select two recent representative systems that provide ablation results with similar prerequisites as our proposed system:
\begin{itemize}[leftmargin=*]
\item VLMnav~\citep{goetting2024endtoend} is an end-to-end VLN system, which assumes no prior knowledge or map; it does not rely on online SLAM for navigation. However, it requires a depth camera to obtain navigability scores (floor mask). It reported $0.210$ SPL on the ObjectNav dataset, its score drops to \uline{$0.136$} without the depth camera.

\item GOAT \citep{chang2023goatthing} is a lifelong mapping system; it requires an RGBD camera to obtain necessary information for navigation and map construction. In addition to the final target, it also takes ``sequential goals'' that help it to take intermediate moves. It reported $0.64$ SPL on the ``in the wild'' dataset with persistent memory, but the score drops to \uline{$0.19$} in the absence of prior knowledge. The authors did not perform an ablation study when depth sensing is removed.
\end{itemize}
In contrast, our proposed work, \textit{VL-Explore}, achieves \uline{$\mathbf{0.357}$} SPL \uline{without requiring} depth sensing, persistent memory, or prior knowledge (\eg~ mapping run) of a scene. To the best of our knowledge, VL-Explore is the \textbf{first VLN system} to achieve efficient indoor visual exploration \textbf{without any depth or range information} (\eg~RGBD, multi-camera, LiDAR).
VL-Explore holds a significant advantage in its \uline{intended use case} -- first-time exploration and target discovery with a single monocular camera.

There also exist many other works based on different prerequisites, which cannot be directly included in quantitative comparison. We instead summarize the characteristics of each system in Table~\ref{tab:vln-comparison}. Compared to existing systems, VL-Explore uniquely requires zero prior knowledge of the task space and operates solely with monocular RGB perception. Additionally, contemporary VLN systems do not directly produce low-level motion commands but instead generate high-level instructions that depend on another traditional localization and navigation pipeline for the eventual motion execution. In contrast, VL-Explore directly outputs low-level motion commands, eliminating the need for auxiliary control and localization systems. This feature makes VL-Explore particularly suitable for rapid deployment on low-cost robots with limited computational and sensing resources.

\subsection{Ablation Study}

\begin{figure}[t]
  \centering
  \vspace{2mm}
  \includegraphics[width=\columnwidth]{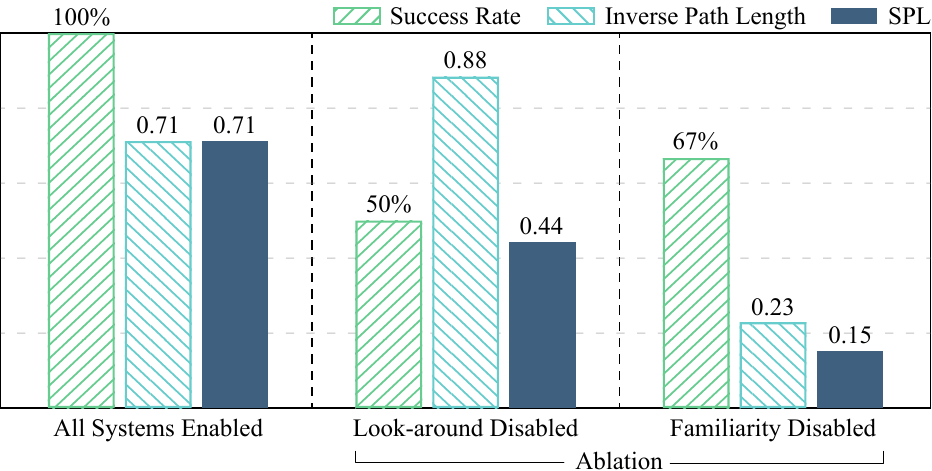} \\
  \vspace{2mm}
  \includegraphics[width=\columnwidth]{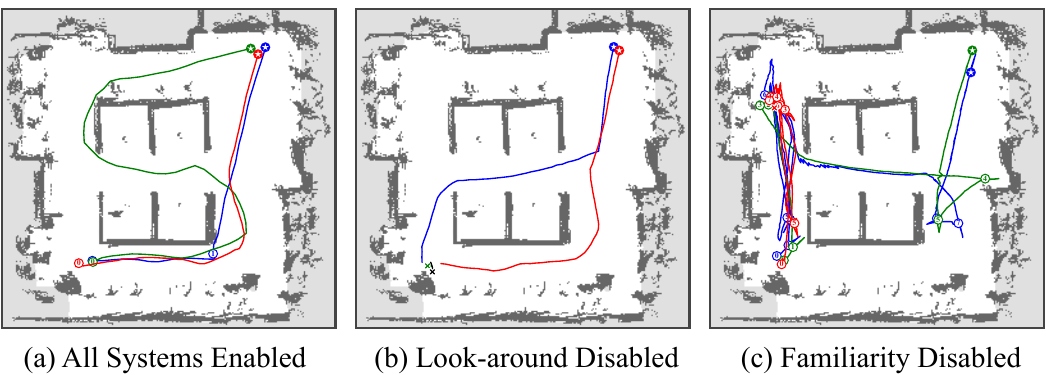}%
  \vspace{-1.5mm}
  \caption{Ablation study of the proposed \BRAND~ system; (a) Comparative results with all systems enabled; (b) Results with the \textit{look-around} subsystem disabled, the robot either completes the task with no turn-around, or directly fail when being trapped, yielding a higher PL but a lower SR; (c) Results with the \textit{familiarity} subsystem disabled, the robot tends to revisit previously explored areas, resulting in longer trajectories and lower success rates.}%
  \label{fig:exp/ablation}
  \vspace{-3mm}
\end{figure}

\newcommand{\NEXTROW}{\\[5pt] \hline \\[-9pt]}

\begin{table*}[t]
    \centering
    \small
    \renewcommand{\arraystretch}{1.2}
    \caption{A qualitative comparison of features and prerequisites among SOTA VLN systems, highlighting the unique advantage of the proposed \BRAND~ system that requires no extra sensor modalities or any separate path-finding or localization steps.}
    \resizebox{\textwidth}{!}{\begin{tabular}{lclllll}
    \Xhline{2\arrayrulewidth}
    \textbf{VLN System} &
    \textbf{Focus} &
    \textbf{Prerequisites} &
    \textbf{Sensing Modality} &
    \textbf{Output Type} &
    \textbf{Deployment} &
    \textbf{Evaluation} \\
    \Xhline{2\arrayrulewidth} \\[-8pt]
    % ==================== CLIP-Nav ====================
    CLIP-Nav~\citep{dorbala2022clipnav} &
    % >> Focus
    Scene &
    % >> Prerequisites
    \Large{$^\text{Pre-captured Images of}_\text{candidate locations}$} &
    % >> In-task Sensor Requirements
    N/A &
    % >> Output Type
    \Large{$^\text{Choice of an Image}_\text{for Next Step}$} &
    % >> Experiment
    Simulation &
    % >> Evaluation Metrics
    \Large{$^\text{Success rate}_\text{(relative)}$}
    \NEXTROW
    % ==================== COW (CLIP on Wheels) ====================
    CLIP on Wheels~\citep{Gadre2022CoWsOP} &
    % >> Focus
    Object &
    % >> Prerequisites
    Unspecified &
    % >> In-task Sensor Requirements
    RGB-D Camera &
    % >> Output Type
    \Large{$^\text{Relevance map of}_\text{the current View}$} &
    % >> Experiment
    Simulation &
    % >> Evaluation Metrics
    \Large{$^\text{Accuracy \&}_\text{Path length}$}
    \NEXTROW
    % ==================== SEEK ====================
    SEEK~\citep{Ginting2024Seek} &
    % >> Focus
    Object &
    % >> Prerequisites
    \Large{$^\text{Pre-built dynamic scene graph}_\text{and relational semantic network}$} &
    % >> In-task Sensor Requirements
    \Large{$^\text{LiDAR, 3 RGB Cameras}_\text{Real-time SLAM}$} &
    % >> Output Type
    \Large{$^\text{Sequence of}_\text{Waypoints}$} &
    % >> Experiment
    % They claimed to have real world experiments, but it's only the "pictures"
    % collected from real world. Not physical robot is used.
    Real World &
    % >> Evaluation Metrics
    \Large{$^\text{Success rate \&}_\text{Path length}$}
    \NEXTROW
    % ==================== PLACEHOLDER ====================
    GCN for Navigation~\citep{kiran2022spatialrelationgraphgraph} &
    % >> Focus
    Object &
    % >> Prerequisites
    \Large{$^\text{Pre-constructed knowledge}_\text{graph of object relations}$} &
    % >> In-task Sensor Requirements
    Monocular RGB Camera &
    % >> Output Type
    \Large{$^\text{Choice of a Pre-}_\text{encoded Action}$} &
    % >> Experiment
    Simulation &
    % >> Evaluation Metrics
    \Large{$^\text{Path length \&}_\text{Success rate}$}
    \NEXTROW
    % ==================== PLACEHOLDER ====================
    Chen~\etal~\citep{Savarese-RSS-19} &
    % >> Focus
    Scene &
    % >> Prerequisites
    \Large{$^\text{Pre-constructed node graph}_\text{of navigable space}$} &
    % >> In-task Sensor Requirements
    N/A &
    % >> Output Type
    Choice of Next Node &
    % >> Experiment
    Simulation &
    % >> Evaluation Metrics
    \Large{$^\text{Success rate \&}_\text{Completion Rate}$}
    \NEXTROW
    % ==================== \BRAND ~ ====================
    \Ours &
    % >> Focus
    Hybrid &
    % >> Prerequisites
    None &
    % >> In-task Sensor Requirements
    Monocular RGB Camera &
    % >> Output Type
    \Large{$^\text{Direct Motion}_\text{Commands (3\textsubscript{DOF})}$} &
    % >> Experiment
    Real World &
    % >> Evaluation Metrics
    \Large{$^\text{Path length \&}_\text{Success rate}$}
    \\ [5pt]
    % ==================== END OF TABLE ====================
    \Xhline{2\arrayrulewidth}
    \end{tabular}}
    \label{tab:vln-comparison}
    \vspace{-2mm}
\end{table*}

In order to understand the contribution of each middleware on the overall performance, we conduct an ablation study by disabling one middleware at a time and re-evaluating the system's performance. The experiment is performed in the medium-sized office scene shown in Fig.~\ref{fig:env}\,(c), with the robot starting from the South-West (SW) corner and tasked to find the target in the North-East (NE) corner.

\begin{itemize}[leftmargin=*]
\item \li{Disabling the \textit{look-around} action.} In this part, the system does not perform any look-around action during a task, including the initial look-around that helps the robot finding its optimal initial heading. The initial orientation of the robot is manually set to a different direction for each trial. The results are shown in Fig.~\ref{fig:exp/ablation}b. With the lack of look-around action, the task would fail upon encountering a dead-end as it could not recover from such a situation. This resulted in a significantly lower success rate ($50\%$). However, in the cases when the robot successfully navigated to the target, the reported inverse path lengths turned out better due to survivorship bias. \ie~inefficient runs tend to fail instead of being recovered by a look-around action.

\item \li{Disabling the \textit{familiarity} middleware.} In this part, the \textit{familiarity} middleware is disabled by truncating the familiarity scores to a fixed value. The results are shown in Fig.~\ref{fig:exp/ablation}c. Without the familiarity middleware, the robot exhibited a strong tendency to revisit previously explored areas. In this particular case, the robot tends to run back and forth between \SW~and \NW~corner of the scene. This lead to significantly longer trajectory lengths (\ie~smaller mean inverse path length), and also induced negative impact on the overall success rate.
\end{itemize}

To summarize, the additional system components added to the VL-Explore pipeline are proven to be essential for its overall performance. In addition to the basic obstacle avoidance and target identification capabilities provided by the vision-language correlation middlewares, the \textit{look-around} action helps the robot to optimize its decision and recover from trap conditions, and the \textit{familiarity} middleware ensures efficient exploration by constantly steering the robot into unexplored area.

% \JI{Check for the mix of present and past tense. It is better to write everything in the present tense. We show, We present, conduct, etc.. Do not mix both}

\section{Limitations, Failure Cases, and Potential Improvements of VL-Explore}

\subsection{Texture-less or Transparent Surfaces}
As a vision-driven system, the proposed VL-Explore\, pipeline encounters challenges in environments with texture-less or transparent surfaces. When presented with a blank or non-informative image, the inherent VLM struggles to generate meaningful responses, disrupting the entire pipeline. For instance, as demonstrated in Fig.~\ref{fig:exp-look-around}\,d (C1), the robot mistakenly classified a water tank with transparent walls as navigable space. This error occurred because the VLM perceived the interior of the tank, visible through its transparent walls, as accessible terrain. To mitigate this issue, a standard deviation threshold was introduced to filter out ambiguous tiles. While this approach improved performance, the system still exhibits a general tendency to misclassify transparent or uniformly colored surfaces, such as blank walls or white floors, as navigable space. These challenges persist, particularly when the robot is positioned close to such surfaces, where the lack of texture further confuses the VLM. Note that this issue with transparent and texture-less walls has been an open problem for 2D indoor navigation by mobile robots~\citep{zhou2017fast}.

%As a pure vision-driven system, our proposed pipeline does not handle well in environments with texture-less or transparent surfaces. When a blank or meaningless image is presented to the VLM, the model will likely fail to generate a meaningful response. This will in turn disrupt the entire pipeline. As shown in Fig.~\ref{fig:exp-look-around}d (C1), the robot falsely classified a water tank with transparent walls as navigable space because it saw through the transparent wall and perceived it's inside as navigable space. The problem has been mitigated with the introduction of the standard deviation threshold on each tile. A general tendency to misclassify transparent surfaces as navigable spaces can still be observed across experiments. This also applies when the robot is too close against a uniformly colored surface, which makes little difference with a white floor to the VLM.

\subsection{Open Space and Artificially Constructed Mazes}
The system's reliance on the semantic understanding capabilities of the VLM limits its effectiveness in large, open spaces. Without sufficient visual cues or meaningful objects in the environment, the robot cannot optimize its exploration based on semantic information. In such cases, the robot defaults to moving straight ahead until it encounters visually significant objects, such as obstacles, at which point its correlation middleware resumes providing meaningful guidance. This limitation reduces the system's efficiency in environments devoid of distinguishing visual features.

%One of the core ideas in the design of our system is to leverage VLM's ability to understand the semantics of a scene. However, this also means that the system is not well equipped to handle open spaces. In an open space, the robot will not be able to optimize it's exploration based on the semantical information provided by the VLM due to lack of visual cues. In this case, the robot will always go straight ahead until it regains visual contact to meaningful objects, \eg~obstacles, and then it's correlation middlewares will start providing meaningful information.

Moreover, our experiments in artificially constructed mazes, such as those built from paper boxes, revealed limitations in the VLM's ability to differentiate between the maze walls, such as monochrome cardboards and floors. As a result, the robot either failed to navigate or attempted to collide with the maze walls. This behavior can be attributed to the lack of similar examples in the VLM’s training data, which likely did not include artificially constructed environments of this nature. Consequently, the system is better suited to real-world scenarios where the VLM has been trained on relevant, diverse visual data. This observation suggests that our proposed system will work effectively in real-world environments where VLMs generally work well.

%Before using the high-bay lab shown in Fig.~\ref{fig:env}, we attempted to build a maze out of paper boxes for testing. However, preliminary experiments turned out unsuccessful because the VLM cannot reliably tell the difference between brown paper box walls apart from white floor. The robot either did not go anywhere or attempted to crush into the boxes. This is understandable since the model's training data does not likely contain many samples of mazes built from paper boxes. Hence it does not generalize well to this scenario.

\subsection{Familiarity Saturation}
The \textit{familiarity} middleware is designed to help the robot avoid revisiting already explored areas by accumulating memory of the environment. However, during extended missions, the familiarity database tends to saturate due to the repetitive nature of objects in the environment. Once all unique objects are recorded, the system struggles to prioritize unexplored areas effectively. Future iterations of the system could implement a memory decay mechanism, allowing older entries in the familiarity database to fade over time, thereby prioritizing new observations and maintaining exploration efficiency.

%The familiarity middleware was designed to accumulate memory of the environment in order to avoid revisiting the same place. However, we observed that the familiarity score tends to saturate in an extended mission. This is caused by the fact that many object in the environment are repetitive, and the familiarity database will eventually pick up all of them. In our current version of implementation, there lacks a mechanism for our familiarity database to gradually fade out older memory in favor of new ones.

\subsection{Adaptive Slicing Strategies}
In our current implementation of VL-Explore, a fixed slicing strategy is used based on the camera frame's aspect ratio, which generally performs well in general. However, in specific scenarios involving narrow navigable spaces (\eg, a $50$\,cm gap between two obstacles), the robot often fails to recognize these spaces as traversable. This is because the narrow passage does not occupy an entire tile, leaving obstacles visible in all tiles and leading the system to avoid the space rather than navigate through it. Future improvements could include the development of advanced slicing strategies that dynamically adapt the \textit{region of interest} based on the surroundings. This could be achieved using sparse object detection or pixel-wise segmentation models, enabling better handling of narrow spaces.

%In the current implementation, the slicing strategy is predetermined based on the aspect ratio of the camera frame. This worked well for simple test environments. However, we observed cases when the robot failed to recognize narrow navigable spaces (\eg a $50\,\text{cm}$ wide space between two paper boxes) because the space is not large enough to fill a tile, leaving visible obstacles in all tiles. This causes the system to avoid the space instead of navigating through it. In future work, a more advanced slicing strategy can be implemented to adaptively foveate its region of interest (ROI) depending on perceived surroundings. This can be achieved by either a bonding-box generation model, or a pixel-wise segmentation model.

\subsection{High Level Reasoning \& Task Narration}
One of the most promising extensions of this work lies in incorporating high-level reasoning and task narration using an integrated LLM. With this addition, the robot could leverage common-sense reasoning to make more intelligent navigation decisions. For example, when prompted to locate an item in a different room, the robot could infer that heading toward a door is the most logical action. Moreover, in bandwidth-limited applications, the LLM could provide real-time task narration by interpreting visual embeddings to describe mission progress. This approach would significantly reduce the communication bandwidth required compared to streaming raw video. By addressing these limitations and pursuing the outlined improvements, the proposed system can evolve into a more robust, efficient, and versatile platform for VLN in complex real-world environments.

%One of the most exciting addition to our system is the ability to understand and reason about the environment at a higher level with the help from a large language model (LLM). For example, the robot can be use common sense to prefer going to a door when it's prompted to find an item in a different room. In addition, in bandwidth limited applications, a remote operator can understand the progress of a mission by using a LLM to narrate the stream of visual embeddings, which requires significantly lower bandwidth than the raw video stream. Our existing architecture already paved the way for such possibilities, but the implementation is left for future work.

\section{Conclusion}
In this paper, we introduced \BRAND, a novel navigation pipeline designed for simultaneous {exploration} and {target discovery} by autonomous ground robots in unknown environments, leveraging the power of VLMs. Unlike traditional approaches that decouple exploration from path planning and rely on inefficient algorithms, \BRAND ~ enables \textbf{zero-shot inference} and efficient navigation using only monocular vision, without prior maps or target-specific information. To validate our approach, we developed a functional prototype UGV platform named \ROVER, optimized for general-purpose VLN tasks in real-world environments. Extensive evaluations demonstrated that \BRAND ~ outperforms state-of-the-art map traversal algorithms in efficiency and achieves comparable performance to path-planning methods that rely on prior knowledge. The integration of a CLIP-based VLM into a real-time navigation system underscores the potential of \BRAND ~ to advance intelligent robotic exploration. As a modular and flexible framework, \BRAND ~ lays the groundwork for future research in applying VLMs to more complex and dynamic robotic applications such as autonomous warehousing, security patrolling, and smart home assistance.

%In this paper, we approached the problem of \textbf{Simultaneous exploration and target discovery} with a novel framework \BRAND, supported by a custom-designed platform \ROVER. We demonstrated the effectiveness of our system in a series of experiments, showing that our proposed system is capable of performing efficient exploration and target discovery tasks in real-world environments. We compared the performance of our system against state-of-the-art algorithms and showed that our system outperforms in terms of efficiency (against map traversal algorithms) and success rate (against path-finding algorithms). The value in our work extends far beyond the immediate results. As a modular and flexible framework, \BRAND will serve as a foundation for future research in embedding vision-language models into \textbf{realtime} robotic systems.

%\section*{Acknowledgments}
%Truncated for the blind review process.
%This work is supported by the NSF grants \# $2330416$.

\section*{Acknowledgments}
This work is supported in part by the National Science Foundation (NSF) grant \#$2330416$; the Office of Naval Research (ONR) grants \#N000142312429, \#N000142312363; and the University of Florida (UF) ROSF research grant \#$132763$

{\small
\bibliographystyle{plainnat} % sort by citation order
\bibliography{refs, refs_vln, robopi_pubs, refs_rebuttal}
}

\appendix

\clearpage
\twocolumn[\section*{Appendix}]
\section{Frame Slicing Strategies}\label{sec:impl/slicing}
% (1) Why use 2x3 tiles, what else has been considered and why they are not used.
To enhance the embedding of positional information within the pipeline, we preprocess the raw camera frames by slicing them into smaller \textit{tiles} before inputting them into the vision-language model. This slicing strategy is specifically designed to optimize the robot's ability to make informed navigation decisions based on its visual perceptions. As illustrated in Fig.~1 in the main paper, the slicer divides each frame into six ($2~\text{rows} \times 3~\text{columns}$) tiles, with the number of horizontal divisions aligned with the robot's control capabilities. Using smaller tiles instead of the full camera frame reduces visual distractions, thereby improving the accuracy of the correlation results.

% (2) Explain that each tile will be processed separately and independently by correlation middlewares, and the dimensions of the correlation scores will correspond to the number of tiles.
The slicing dimensions are consistently maintained across all stages of the pipeline. For instance, under the $2\times3$  slicing strategy, a frame divided into six tiles produces a corresponding 
$2\times3\times512$ matrix of visual embeddings. These embeddings are subsequently aggregated and passed to the correlation middleware, which generates a $2\times3$ matrix of correlation scores, with each element corresponding to a specific tile. The influence of individual tiles on one another is determined by the logic of the respective correlation middleware. For example, while the familiarity middleware disallows correlations between tiles within the same frame, it may permit correlations between the lower-left tile of a previous frame and the upper-right tile of the current frame.

%The slicing dimensions are preserved across all parts of the pipeline. Take the $2 \times 3$ slicing strategy as an example, a $2 \times 3$ matrix of sliced tiles will produce a $2 \times 3 \times 512$ matrix of visual embeddings. These embeddings are then packed together and passed down to the correlation middleware. Each correlation middleware will then produce a $2 \times 3$ matrix of correlation scores, where each element corresponds to a tile in the sliced frame. How different tiles affect each other is determined by each correlation middleware. For example, the familiarity middleware does not permit correlation between tiles within the same frame, but it might correlate the lower left tile from a previous frame to the upper right tile on the current frame.

% (3) Explain how each tiles are slightly "expanded" to create overlaps between them.
To preserve spatial context in each tile, sliced regions are slightly expanded beyond their original boundaries. This overlap introduces approximately $20\%$ shared area between adjacent tiles, mitigating the loss of contextual information. This adjustment is beneficial when the camera encounters texture-less surfaces, such as walls, paper boxes, or door panels -- where the absence of distinctive features could otherwise impair the robot's perception and navigation performance.

\section{Text Prompt Generation}
We also design a prompt system to compose text prompts in a hierarchical template. The proposed design is inspired by WinCLIP~\citep{Jeong2023CVPR,zhu2024toward}, an anomaly detection framework. Our prompt system consists of the following templates and notations, which we follow throughout the paper.
\begin{itemize}[leftmargin=*]
\item \li{Top-level prompts} such as:
\begin{itemize}[leftmargin=*]
    \item \Prompt{A \DscPrompt{desc.} photo of a \SttPrompt{state} \ObjPrompt{object}}
    \item \Prompt{A \DscPrompt{desc.} image of a \SttPrompt{state} \ObjPrompt{object}}
    \item \Prompt{A \SttPrompt{state} \ObjPrompt{object}}
\end{itemize}
\item \li{\DscPrompt{descriptions}} such as \ttt{clear}, \ttt{blurry}, \ttt{cropped}, \ttt{empty}, \ttt{corrupted}, \etc
\vspace{1mm}
\item \li{\SttPrompt{states}} such as \ttt{clean}, \ttt{clear}, \ttt{wide}, \ttt{narrow}, \ttt{cluttered}, \ttt{messy}, \ttt{way blocking}, \etc
\vspace{1mm}
\item \li{\ObjPrompt{objects}} such as \ttt{floor}, \ttt{wall}, \ttt{door}, \ttt{object}, \etc
\end{itemize}
%\noindent As is shown above, we use different types of brackets to indicate different types of template substitutions. All figures in this paper follows the same convention.

We utilize a YAML-based syntax to define the structure of the prompt databases. YAML, a widely adopted and human-readable markup language, has been extended in our implementation to incorporate additional syntactical features for enhanced flexibility and fine-grained control over the generation of text prompt combinations. Two key features introduced are \textit{selective integration} and \textit{in-place expansion}. 
\begin{enumerate}[label={$\arabic*$.},nolistsep,leftmargin=*]
\item  \li{Selective integration} enables a prompt to bypass certain levels of the template hierarchy, thereby allowing precise customization. For instance, a prompt such as \Prompt{A \SttPrompt{meaningless} photo} should not be followed by any \ObjPrompt{object}. This is achieved by terminating the prompt early, omitting the insertion marker (\ie~\ttt{\footnotesize{\{\}}}) in its definition. 

\item \li{In-place expansion} facilitates the collective definition of similar short prompts by using a vertical bar syntax (\ie~\ttt{|}) to separate terms. This feature is particularly effective when used in conjunction with \textit{selective integration}. For example, the prompt \Prompt{A photo with no \ObjPrompt{context|texture|information}} is expanded into three distinct prompts. Although not fitting into the hierarchical template structure, these can be defined as top-level negative prompts, aiding in distinguishing meaningful content from irrelevant or meaningless information. 
\end{enumerate}
These enhancements to the YAML-based syntax provide an effective framework for defining and managing prompt databases with high precision and adaptability.

%We use a {\tt YAML} based syntax to define the \textit{prompt databases}. {\tt YAML} is a popular human-friendly mark-up language. Our implementation features additional syntaxes that allows for more flexible and find-grained control over the generated combinations of text prompts: \li{Selective Integration} allows a prompt to skip integrating certain levels of the template. For example, prompt \Prompt{A \SttPrompt{meaningless} photo} should not be followed by any \ObjPrompt{object}. A prompt can be terminated early by excluding the insertion mark (\ie~\ttt{\footnotesize{\{\}}}) from its definition. \li{In-place expansion} allows for similar short prompts to be defined collectively. The syntax applies to consecutive terms divided by vertical bars (\ie \ttt{|}). It is particularly useful when used alongside selective integration. For example, prompt \Prompt{A photo with no \ObjPrompt{context|texture|information}} will be expanded into three separate prompts. These prompts would not fit into the template hierarchy, but thanks to the aforementioned features, they can be defined as top-level negative prompt to help differentiate useful information from meaningless ones.

\vspace{-1mm}
\section{Performance Optimization}
Navigation decisions are a critical component of real-time autonomous robotic systems, requiring rapid execution to ensure safety and operational efficiency. However, most vision-language models are designed as extensions of LLMs, where processing delays and data throughput are less critical. To evaluate the suitability of VLMs for robotic navigation, we identify two key performance metrics:
%Navigation decision is a critical component of a real-time autonomous robotic system, requiring rapid execution. In contrast, most vision-language models are designed as extensions of large language models, where delays and data throughput are less critical. Therefore, we identify two key metrics that reflect the performance of a VLM-based navigation system:
\begin{enumerate}[label={$\arabic*$.},nolistsep,leftmargin=*]
\item  \li{Decision delay} measures the time elapsed between capturing a frame from the robot’s camera and issuing a corresponding motion command to the motors. This metric reflects the system's ability to promptly react to environmental changes, such as avoiding obstacles and maintaining safe navigation. 
\item \li{Throughput} quantifies the number of frames processed per second, directly influencing the smoothness of the robot’s motion. Limited computational resources necessitate dropping unprocessed frames, retaining only the latest frame for decision-making. Higher throughput minimizes inter-frame discrepancies, reducing abrupt changes in motion and ensuring smoother transitions. Mathematically, throughput is the inverse of the decision delay: \ie~$f=\frac{1}{T_{\text{delay}}}$.
\end{enumerate}

\vspace{1mm}
\noindent
On the other hand, most robotic systems rely on embedded computers featuring RISC architecture processors and power-limited batteries, prioritizing power efficiency over computational capability. Addressing these constraints, we developed an optimized architecture to maximize resource utilization while improving decision delay and throughput.

Focusing on reducing decision delay and increasing data throughput, our proposed framework divides the navigation pipeline into four major nodes: pre-processing, inference, correlation, and decision-making. Computationally intensive tasks are offloaded to the GPU to exploit parallel processing capabilities. These optimizations were implemented on a power-limited embedded system, balancing efficiency and performance. With the proposed optimizations, we achieve: (\textbf{i}) decision delay: $252.10$\,ms, a $900\%$ improvement; and (\textbf{ii}) throughput: $5.01$ FPS (frames per second), representing a $400\%$ improvement compared to the sequential CPU-based implementation. These results, detailed in Table \ref{tab:perf}, demonstrate the viability of our VLM-based navigation pipeline for real-time robotic applications under computational constraints.

\begin{table}[t]
    \centering
    \renewcommand{\arraystretch}{1.1}
    \newcommand{\dg}{\textsuperscript{\textdagger}}
    \caption{Performance margins of the navigation pipeline of \BRAND ~ at different optimization levels. The acronyms: \acronym{P} Preprocess,  \acronym{I} Inference, \acronym{C} Correlation, and \acronym{D} Decision.}%
    \vspace{-1mm}
    \resizebox{\columnwidth}{!}{
    \begin{tabular}{c|c|c|c|c|c|c}
    \Xhline{2\arrayrulewidth}
    \multirow{2}{*}{\small{Config.}} & \multicolumn{5}{c|}{\small{Mean Data Processing Delay (ms)}} & \multirow{2}{*}{\Large{$^\text{Throughput}_\text{(mean FPS)}$}} \\ \cline{2-6}
                    & \acronym{P} & \acronym{I} & \acronym{C} & \acronym{D} & \small{Total} & \\ \hline
    CPU Seq.\,& $80.57$ & $ 2221.67$ & $\,3.13\,$ & $\,0.37\,$ & $ 2534.55$ & $0.41\, \pm 0.07$ \\ \hline
    CPU Para. & $73.16$ & $ 2201.84$ & $\,3.29\,$ & $\,0.40\,$ & $ 2463.26$ & $0.44\, \pm 0.02$ \\ \hline
    GPU Seq.\,& $65.93$ & $\z178.77$ & $\,2.94\,$ & $\,0.45\,$ & $\z327.57$ & $3.22\, \pm 0.25$ \\ \hline
    GPU Para. & $25.32$ & $\z171.08$ & $\,3.43\,$ & $\,0.48\,$ & $\z252.10$ & $\mathbf{5.01} \pm 0.36$ \\
    \Xhline{2\arrayrulewidth}
    \multicolumn{7}{l}{}\\[-4mm] % Add some space between the table and the footnote
    % \multicolumn{7}{l}{\small{
    %     \textbf{\textit{Acronyms}}:
    %     \acronym{P} Preprocess,  \acronym{I} Inference,
    %     \acronym{C} Correlation, \acronym{D} Decision.
    % }} \\
    % \multicolumn{7}{l}{\small{
    %     \dg~Showing typical (mean) values sampled across multiple iterations.
    % }}
    \end{tabular}
    }
    \label{tab:perf}
    \renewcommand{\arraystretch}{1.0}
    \vspace{-3mm}
\end{table}

%We developed an architecture that utilizes as many resources as possible on such a platform to improve both decision delay and data throughput. Four major nodes (preprocess, inference, correlation and decision) are divided into separate threads and are executed in parallel. All computationally intensive tasks are offloaded to the GPU. With there optimizations in place, we achieved a data processing delay of $252.10$ ms and a throughput of $5.37$. As shown in Table. \ref{tab:perf}, that is $800\%$ improvement for the delay and $400\%$ improvement for the throughput compared to the sequential CPU implementation.

\section{Usage of 2D LiDAR}\label{sec:LiDAR}
The proposed \BRAND ~ pipeline uses only monocular RGB data for VLN navigation decisions. Nevertheless, we installed a 2D $360$\degree\ scanning LiDAR on the robot for experimental safety, map generation, and comparative analyses. The specific LiDAR functions are listed below.
\begin{enumerate}[label={$\arabic*$.},nolistsep,leftmargin=*]
\item  \li{Emulation of a proximity kill-switch.} The LiDAR serves as a virtual \textit{kill switch} during operation, detecting obstacles in the robot’s intended path. It monitors the motion commands sent to the wheels to infer the robot’s planned trajectory and checks for potential obstructions along that direction.  A {\textit{halt signal}} is asserted upon detection of a potential collision, thus strictly used for safety purposes without influencing the navigation algorithm. 

\item  \li{Visualization of map and robot trajectory.} Mapping is performed offline using recorded LiDAR and odometry data, with subsequent trajectory analysis to evaluate navigation efficiency and overall performance. We utilize the \textit{SLAM Toolbox}~\citep{Macenski2021} to generate maps and trajectories, implementing a two-pass process to enhance the quality of the results. In the first pass, the SLAM Toolbox operates in offline mapping mode, creating a high-resolution 2D map of the environment. In the second pass, the pre-constructed map is loaded and run in localization-only mode to generate accurate trajectories using LiDAR localization.
Despite this two-pass process, the trajectories generated often exhibit zig-zag patterns caused by a mismatch between SLAM drift correction frequency and the robot's odometry update rate. These patterns are not observed in the actual motion of the robot; still, they affect our performance evaluation due to increased lengths of the projected paths. To mitigate this, we introduce a \textit{stride} parameter to smooth the trajectory. This smoothing parameter reduces the projected trajectory length and better matches the robot's actual travel distance.

%The mapping is performed offline using recorded data, and the resulting trajectory is analyzed to assess the robot's navigation efficiency and overall performance. We employ \textit{slam toolbox} to generate maps and trajectories from recorded LiDAR and odometry data. We designed a 2-pass process to help improve the quality of results. In the 1st pass, \textit{slam toolbox} was ran in offline mapping mode, creating a high-quality 2D map of the environment. In the 2nd pass, \textit{slam toolbox} was configured to load back the map constructed from the 1st pass and run itself in localization-only mode. This pass generates an accurate trajectory derived from LiDAR localization. All maps and trajectories shown in this paper were generated using the aforementioned process. It is worth noting that, even with the 2-pass process, the generated trajectories still contain zig-zag patterns due to mismatch between SLAM drift correction frequency and robot's odometry update frequency. As is shown in Fig.~\ref{fig:trj-stride}a, such patterns were not observed in the actual motion of the robot. These patterns negatively affect our evaluation of the robot's navigation performance as it increases the projected path length. To mitigate this issue, we introduce a \textit{stride} parameter to smooth the trajectory. The effect of this parameter is demonstrated in Fig.~\ref{fig:trj-stride}b. The total length (``travel'') of the sample trajectory reduced from $8.53~\text{m}$ to $5.11~\text{m}$, which is considered closer to the ground truth.
\item \li{Comparison with simulated algorithms.} A 2D map generated using LiDAR data is employed to simulate and evaluate traditional range sensor-based map traversal and path-planning algorithms for performance comparison. The LiDAR data collected during \BRAND ~ experiments was used to reconstruct the exploration map, which is reused for simulations. This approach ensured that all comparisons were conducted on the same map, providing a fair and consistent evaluation of the different algorithms.
\end{enumerate}

\section{Simulation Algorithms}

\vspace{1mm}
\noindent
For map traversal algorithms~\citep{pang2019randoomwalk}, we consider:
\begin{enumerate}[label={$\arabic*$.},nolistsep,leftmargin=*]
\item  \li{Random Walk:} Starts with a given heading, then randomly selects a new heading when an obstacle is encountered.
\item  \li{Wall Bounce:} Starts at a given heading, then bounces off obstacles based on the normal vector of the impact point.
\item  \li{Wave Front:} Starts as a Gaussian probability distribution, then spreads outwards and bounces off obstacles. A detailed explanation is provided below.
\end{enumerate}

\vspace{1mm}
\noindent
Additionally, we compare the following Bug Algorithms~\citep{lumelsky1986dynamic,lumelsky1987path}:
\begin{enumerate}[label={$\arabic*$.},nolistsep,leftmargin=*]
\item  \li{Bug0:} Moves straight toward the target until encountering an obstacle, then follows the obstacle boundary until it can resume a direct path to the target.
\item  \li{Bug1:} Heads towards the target when possible, otherwise, circumnavigates the obstacle. After looping an obstacle, travels to the point with the minimum distance on the loop.
\item  \li{Bug2:} Circumnavigates obstacles upon encountering until it crosses the direct line from the start to the target (the {\tt m-line}), then resumes a straight path toward the target.
\end{enumerate}

\subsection{Wave Front Simulation}

The ``Wave Front'' simulation is a custom-developed package to serve as a general baseline metric for comparison between different methods. It utilizes the concept of classic wave function~\citep{Maxwell1865} and interprets it as a probability distribution, allowing simultaneous exploration of infinitely many directions driven by the following equation:
\begin{equation}
\frac{\partial^2}{\partial t^2}\,\psi(x, y, t) = c^2 \nabla^2\psi(x, y, t)
\end{equation}
The simulation implements a 2D Laplacian equation at discretized time steps. Given wave speed $c$, time step $\Delta t$, and spatial step $\Delta x$, the wave equation is discretized as:
\begin{align*}
\psi(x, y, t_{n+1})
=~& 2\cdot \psi(x, y, t_{n}) - \psi(x, y, t_{n-1}) \\
+~& \z~ ~ \frac{c^2 \Delta t^2}{\Delta x^2} \nabla^2\psi(x, y, t) \\
s.t.~& \z~ ~ t_{n+1} - t_{n} = \Delta t
\end{align*}
Here, the discretized Laplacian operator $\nabla^2$~\citep{FDTD2000Wave} is defined as:
\begin{align*}
\nabla^2\psi(x, y)
& = \psi(x_{i+1},\,y_j)
  + \psi(x_{i-1},\,y_j) \\
& + \psi(x_i,\,y_{j+1})
  + \psi(x_i,\,y_{j-1}) \\
& - 4\cdot\psi(x, y) \\
s.t.~& ~ ~ ~ \, x_{i+1} - x_i = y_{j+1} - y_j = \Delta x
\end{align*}
Besides, map boundaries and obstacles are simulated by reflective conditions. That is, at boundary point $(x, y)$, we have: $\nabla\psi(x, y) \cdot \mathbf{\hat{n}}_{x, y} = 0$, where $\mathbf{\hat{n}}_{x, y}$ is the normal vector.
In a discretized 2D grid, the directions of this normal vector are simplified to four possibilities, \ie, $\mathbf{\hat{n}} \in \left\{ \pm \mathbf{\hat{x}}, \pm \mathbf{\hat{y}} \right\}$.

To implement the aforementioned boundary condition, the discretized Laplacian operator can be rewritten as:
\begin{align*}
    \nabla^2\psi(x,\,y)
    & = k_{i+1,\,j} \cdot \psi_{i+1,\,j}
      + k_{i-1,\,j} \cdot \psi_{i-1,\,j} \\
    & + k_{i,\,j+1} \cdot \psi_{i,\,j+1}
      + k_{i,\,j-1} \cdot \psi_{i,\,j-1} \\
    & - (k_{i+1,\,j} + k_{i-1,\,j} + k_{i,\,j+1} + k_{i,\,j-1})\cdot\psi(x,\,y)
\end{align*}
{\small\begin{equation*}
s.t.~ ~ ~ ~ \, k_{i,j} = \left\{~\begin{matrix}
        1 & \text{if } (x_i,\,y_j) \text{ is free space} \\[2mm]
        0 & \text{if } (x_i,\,y_j) \text{ is obstacle ~} \\
    \end{matrix}\right.
\end{equation*}}

At time $T = 0$, a probability distribution is initialized as a Gaussian distribution centered at the robot's starting location, with a standard deviation set to half of the robot's size; the total probability is normalized to $1$. At each iteration, the wave function is multiplied by an inverse Gaussian function centered around the target location, effectively reducing the total probability on the map and simulating a draining effect. The simulation terminates when the total remaining probability falls below a defined threshold ($1e^{-4}$ in our experiments). The outcome is a 2D probability distribution $p(t)$ indicating the likelihood of the robot reaching the target at a given time point. The mean and standard deviation of this distribution are used for comparison with other trajectory-based algorithms.

At a high level, this algorithm simulates the behavior of infinitely many wall-bouncing robots moving in all possible directions. Notably, the drain operation creates a uniform gradient toward the target, subtly guiding the robot. As a result, the algorithm performs slightly better than a purely randomized wall-bouncing simulation, which lacks any target information. The travel distance when the first non-zero probability is detected is used as the baseline distance for each origin-target combination. For relative performance comparisons, the travel distances are normalized by this baseline distance before aggregating the results.

\section{Additional Experimental Results}

\begin{figure*}
    \centering
    \includegraphics[width=\textwidth]{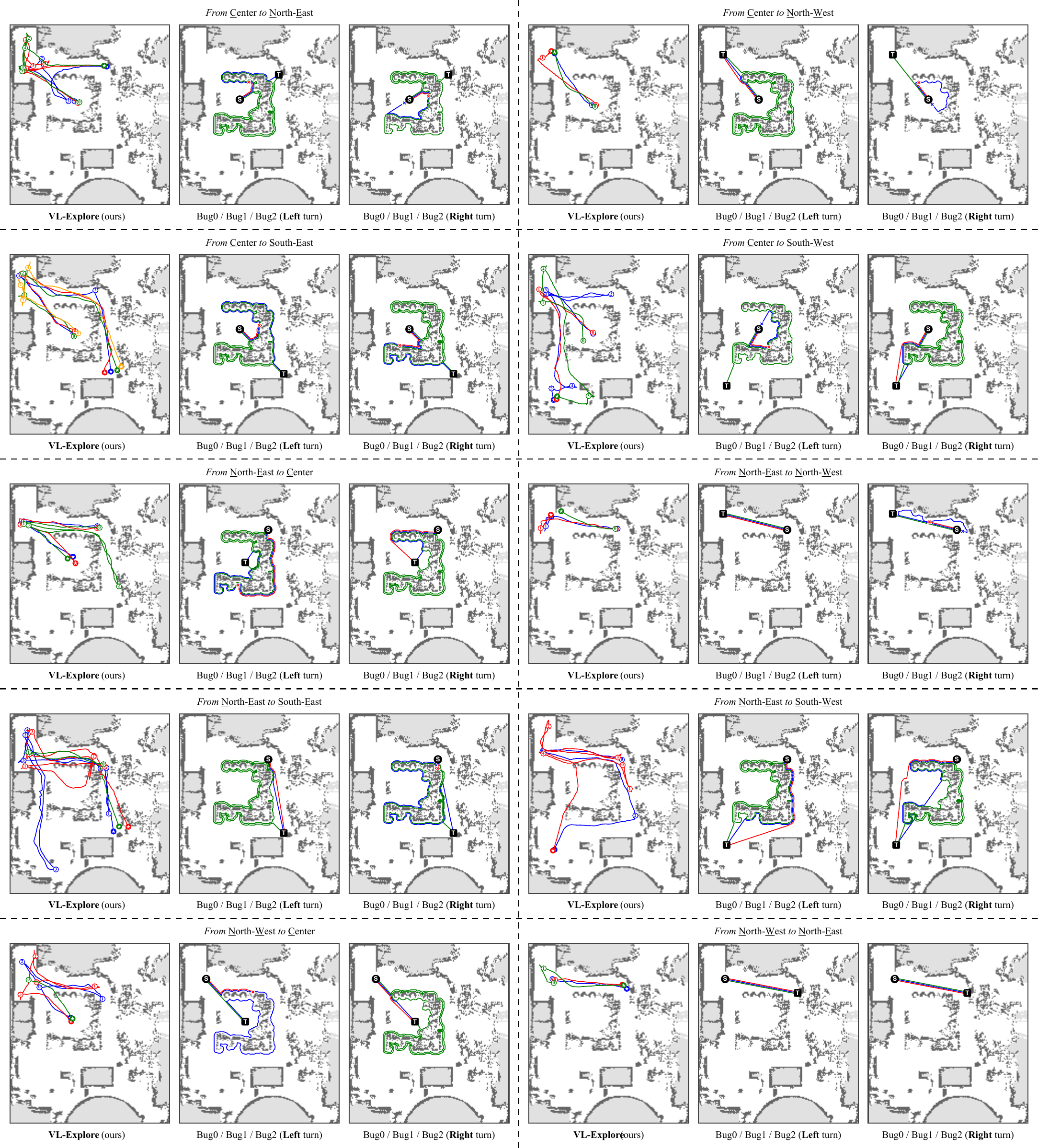}
    \caption{
        Trajectories of all results reported in the experiment section (part.1)
    }
    \label{fig:trajectories-1}
\end{figure*}

\begin{figure*}
    \centering
    \includegraphics[width=\textwidth]{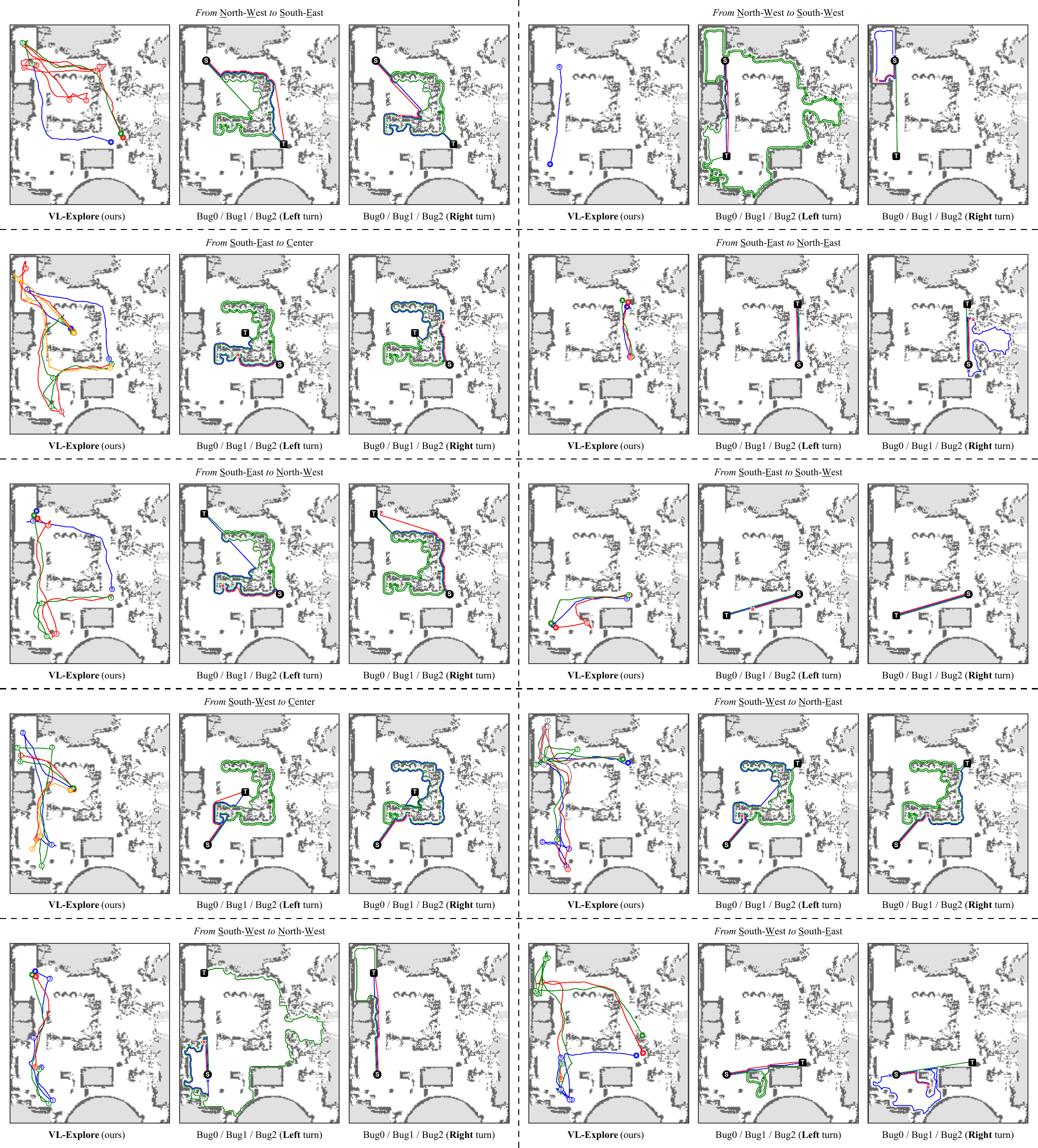}
    \caption{
        Trajectories of all results reported in the experiment section (part.2)
    }
    \label{fig:trajectories-2}
\end{figure*}

The full set of trajectories recorded in the experiment is shown in Fig.~\ref{fig:trajectories-1} and Fig.~\ref{fig:trajectories-2}. Metrics measured for each source-target combination are shown in Table~\ref{tab:results}.

\newcommand\Failed{{\color{red}$\times$}}

\begin{table*}
    \centering
    \small
    \renewcommand{\arraystretch}{1.2}
    \caption{Quantitative performance comparison of the exploration (random walk, wall bounce, and wavefront) and path-finding (Bug 0/1/2) algorithms in terms of trajectory length. All units are in meters; see Fig.~\ref{fig:env} for the source-target layouts. The acronyms: \acronym{L} and \acronym{R} represent left and right turn rules, respectively; the failure cases are shown as `\Failed' markers.}
    \begin{tabular}{cccccccccccc}
    \Xhline{2\arrayrulewidth}
    \multicolumn{2}{c}{Task Setup} &
    \Ours &
    Random\,Walk &
    Wall\,Bounce &
    Wave\,Front &
    \multicolumn{2}{c}{Bug0} &
    \multicolumn{2}{c}{Bug1} &
    \multicolumn{2}{c}{Bug2} \\ \cline{1-12}
    Source & Target &
    Avg.\,\footnotesize{$\pm$ Std.} &
    Avg.\,\footnotesize{$\pm$ Std.} &
    Avg.\,\footnotesize{$\pm$ Std.} &
    Avg.\,\footnotesize{$\pm$ Std.} &
    \acronym{L} & \acronym{R} &
    \acronym{L} & \acronym{R} &
    \acronym{L} & \acronym{R} \\
    \Xhline{2\arrayrulewidth}
    \multicolumn{12}{c}{\textbf{Warehouse Scene ($16m \times 18m$)}} \\
    \Xhline{2\arrayrulewidth}
    \multirow{4}{*}{\C}
    & \NW &
        $\z8.3$\footnotesize{$\pm \z2.1$} &  % \BRAND
        $149.6$\footnotesize{$\pm  18.6$} &  % RandomWalk
        $169.3$\footnotesize{$\pm  14.9$} &  % WallBounce
        $162.7$\footnotesize{$\pm  55.8$} &  % WaveFront
        $\z5.9$                           &  % Bug0L
        \Failed                           &  % Bug0R
        $ 75.8$                           &  % Bug1L
        $ 42.2$                           &  % Bug1R
        $\z6.0$                           &  % Bug2L
        \Failed                           \\ % Bug2R
        \cline{2-12}
    & \NE &
        $ 27.2$\footnotesize{$\pm \z2.7$} &  % \BRAND
        $326.5$\footnotesize{$\pm  27.0$} &  % RandomWalk
        $313.1$\footnotesize{$\pm  19.4$} &  % WallBounce
        $247.4$\footnotesize{$\pm  74.3$} &  % WaveFront
        \Failed                           &  % Bug0L
        \Failed                           &  % Bug0R
        $ 63.0$                           &  % Bug1L
        $ 49.5$                           &  % Bug1R
        $ 13.4$                           &  % Bug2L
        \Failed                           \\ % Bug2R
        \cline{2-12}
    & \SW &
        $ 29.0$\footnotesize{$\pm \z7.1$} &  % \BRAND
        $256.7$\footnotesize{$\pm  19.4$} &  % RandomWalk
        $233.4$\footnotesize{$\pm  15.6$} &  % WallBounce
        $236.9$\footnotesize{$\pm  66.8$} &  % WaveFront
        \Failed                           &  % Bug0L
        $\z7.9$                           &  % Bug0R
        $ 43.7$                           &  % Bug1L
        $ 72.5$                           &  % Bug1R
        \Failed                           &  % Bug2L
        $\z8.6$                           \\ % Bug2R
        \cline{2-12}
    & \SE &
        $ 27.0$\footnotesize{$\pm \z3.7$} &  % \BRAND
        $370.6$\footnotesize{$\pm  23.4$} &  % RandomWalk
        $338.4$\footnotesize{$\pm  18.8$} &  % WallBounce
        $285.2$\footnotesize{$\pm  82.5$} &  % WaveFront
        \Failed                           &  % Bug0L
        \Failed                           &  % Bug0R
        $ 54.9$                           &  % Bug1L
        $ 59.2$                           &  % Bug1R
        $ 23.2$                           &  % Bug2L
        $ 19.4$                           \\ % Bug2R
        \Xhline{2\arrayrulewidth}
    \multirow{4}{*}{\NW}
    & \C &
        $ 21.6$\footnotesize{$\pm  10.9$} &  % \BRAND
        $233.3$\footnotesize{$\pm  26.0$} &  % RandomWalk
        $193.6$\footnotesize{$\pm  16.2$} &  % WallBounce
        $192.2$\footnotesize{$\pm  55.8$} &  % WaveFront
        \Failed                           &  % Bug0L
        $\z5.9$                           &  % Bug0R
        $ 47.0$                           &  % Bug1L
        $ 67.7$                           &  % Bug1R
        $ 40.1$                           &  % Bug2L
        $\z6.0$                           \\ % Bug2R
        \cline{2-12}
    & \NE &
        $\z8.7$\footnotesize{$\pm \z2.2$} &  % \BRAND
        $253.7$\footnotesize{$\pm  25.4$} &  % RandomWalk
        $220.1$\footnotesize{$\pm  18.3$} &  % WallBounce
        $233.5$\footnotesize{$\pm  61.0$} &  % WaveFront
        $\z7.4$                           &  % Bug0L
        $\z7.4$                           &  % Bug0R
        $\z7.4$                           &  % Bug1L
        $\z7.4$                           &  % Bug1R
        $\z7.4$                           &  % Bug2L
        $\z7.4$                           \\ % Bug2R
        \cline{2-12}
    & \SW &
        $\z9.9$\footnotesize{$\pm \z0.0$} &  % \BRAND
        $290.2$\footnotesize{$\pm  21.7$} &  % RandomWalk
        $241.5$\footnotesize{$\pm  14.6$} &  % WallBounce
        $239.4$\footnotesize{$\pm  69.0$} &  % WaveFront
        $\z9.6$                           &  % Bug0L
        \Failed                           &  % Bug0R
        $132.1$                           &  % Bug1L
        $ 84.4$                           &  % Bug1R
        $ 10.0$                           &  % Bug2L
        \Failed                           \\ % Bug2R
        \cline{2-12}
    & \SE &
        $ 29.6$\footnotesize{$\pm  17.9$} &  % \BRAND
        $338.8$\footnotesize{$\pm  24.9$} &  % RandomWalk
        $332.7$\footnotesize{$\pm  19.9$} &  % WallBounce
        $281.5$\footnotesize{$\pm  78.7$} &  % WaveFront
        $ 14.8$                           &  % Bug0L
        \Failed                           &  % Bug0R
        $ 63.8$                           &  % Bug1L
        $ 52.0$                           &  % Bug1R
        $ 15.7$                           &  % Bug2L
        $ 24.8$                           \\ % Bug2R
        \Xhline{2\arrayrulewidth}
    \multirow{4}{*}{\NE}
    & \C &
        $ 23.6$\footnotesize{$\pm  12.1$} &  % \BRAND
        $275.6$\footnotesize{$\pm  24.5$} &  % RandomWalk
        $251.2$\footnotesize{$\pm  20.0$} &  % WallBounce
        $223.4$\footnotesize{$\pm  62.6$} &  % WaveFront
        \Failed                           &  % Bug0L
        $\z9.2$                           &  % Bug0R
        $ 50.5$                           &  % Bug1L
        $ 60.9$                           &  % Bug1R
        $ 28.0$                           &  % Bug2L
        $ 12.0$                           \\ % Bug2R
        \cline{2-12}
    & \NW &
        $\z9.1$\footnotesize{$\pm \z2.4$} &  % \BRAND
        $157.5$\footnotesize{$\pm  21.5$} &  % RandomWalk
        $187.0$\footnotesize{$\pm  17.0$} &  % WallBounce
        $181.9$\footnotesize{$\pm  45.6$} &  % WaveFront
        $\z6.9$                           &  % Bug0L
        \Failed                           &  % Bug0R
        $\z6.9$                           &  % Bug1L
        $ 75.7$                           &  % Bug1R
        $\z6.9$                           &  % Bug2L
        \Failed                           \\ % Bug2R
        \cline{2-12}
    & \SW &
        $ 42.0$\footnotesize{$\pm \z9.5$} &  % \BRAND
        $236.2$\footnotesize{$\pm  19.3$} &  % RandomWalk
        $250.9$\footnotesize{$\pm  14.2$} &  % WallBounce
        $237.9$\footnotesize{$\pm  70.5$} &  % WaveFront
        $ 14.3$                           &  % Bug0L
        $ 13.9$                           &  % Bug0R
        $ 58.9$                           &  % Bug1L
        $ 55.2$                           &  % Bug1R
        $ 17.1$                           &  % Bug2L
        $ 24.0$                           \\ % Bug2R
        \cline{2-12}
    & \SE &
        $ 48.7$\footnotesize{$\pm  19.4$} &  % \BRAND
        $235.8$\footnotesize{$\pm  22.5$} &  % RandomWalk
        $277.9$\footnotesize{$\pm  18.2$} &  % WallBounce
        $240.1$\footnotesize{$\pm  67.9$} &  % WaveFront
        $\z7.6$                           &  % Bug0L
        \Failed                           &  % Bug0R
        $ 68.3$                           &  % Bug1L
        $ 44.7$                           &  % Bug1R
        $\z7.7$                           &  % Bug2L
        $ 41.0$                           \\ % Bug2R
        \Xhline{2\arrayrulewidth}
    \multirow{4}{*}{\SW}
    & \C &
        $ 19.3$\footnotesize{$\pm \z6.0$} &  % \BRAND
        $340.9$\footnotesize{$\pm  26.8$} &  % RandomWalk
        $331.7$\footnotesize{$\pm  19.1$} &  % WallBounce
        $266.0$\footnotesize{$\pm  66.2$} &  % WaveFront
        \Failed                           &  % Bug0L
        \Failed                           &  % Bug0R
        $ 66.1$                           &  % Bug1L
        $ 50.0$                           &  % Bug1R
        $ 12.0$                           &  % Bug2L
        $ 33.7$                           \\ % Bug2R
        \cline{2-12}
    & \NW &
        $ 15.9$\footnotesize{$\pm \z4.5$} &  % \BRAND
        $329.3$\footnotesize{$\pm  22.6$} &  % RandomWalk
        $330.6$\footnotesize{$\pm  17.2$} &  % WallBounce
        $259.9$\footnotesize{$\pm  63.8$} &  % WaveFront
        \Failed                           &  % Bug0L
        $ 10.5$                           &  % Bug0R
        $ 65.1$                           &  % Bug1L
        $ 19.7$                           &  % Bug1R
        \Failed                           &  % Bug2L
        $ 10.9$                           \\ % Bug2R
        \cline{2-12}
    & \NE &
        $ 31.1$\footnotesize{$\pm \z3.1$} &  % \BRAND
        $396.7$\footnotesize{$\pm  29.0$} &  % RandomWalk
        $288.3$\footnotesize{$\pm  16.7$} &  % WallBounce
        $276.8$\footnotesize{$\pm  78.6$} &  % WaveFront
        \Failed                           &  % Bug0L
        \Failed                           &  % Bug0R
        $ 51.8$                           &  % Bug1L
        $ 65.1$                           &  % Bug1R
        $ 27.5$                           &  % Bug2L
        $ 15.8$                           \\ % Bug2R
        \cline{2-12}
    & \SE &
        $ 31.8$\footnotesize{$\pm  11.2$} &  % \BRAND
        $217.2$\footnotesize{$\pm  22.0$} &  % RandomWalk
        $208.3$\footnotesize{$\pm  16.5$} &  % WallBounce
        $220.7$\footnotesize{$\pm  62.3$} &  % WaveFront
        $\z7.8$                           &  % Bug0L
        \Failed                           &  % Bug0R
        $ 10.1$                           &  % Bug1L
        $ 71.4$                           &  % Bug1R
        $\z8.0$                           &  % Bug2L
        \Failed                           \\ % Bug2R
        \Xhline{2\arrayrulewidth}
    \multirow{4}{*}{\SE}
    & \C &
        $ 29.7$\footnotesize{$\pm \z8.0$} &  % \BRAND
        $386.0$\footnotesize{$\pm  25.9$} &  % RandomWalk
        $311.1$\footnotesize{$\pm  17.7$} &  % WallBounce
        $283.3$\footnotesize{$\pm  72.9$} &  % WaveFront
        \Failed                           &  % Bug0L
        $ 16.1$                           &  % Bug0R
        $ 57.2$                           &  % Bug1L
        $ 54.5$                           &  % Bug1R
        $ 18.2$                           &  % Bug2L
        $ 21.2$                           \\ % Bug2R
        \cline{2-12}
    & \NW &
        $ 22.5$\footnotesize{$\pm \z4.4$} &  % \BRAND
        $369.9$\footnotesize{$\pm  26.5$} &  % RandomWalk
        $298.7$\footnotesize{$\pm  17.7$} &  % WallBounce
        $262.8$\footnotesize{$\pm  63.5$} &  % WaveFront
        \Failed                           &  % Bug0L
        \Failed                           &  % Bug0R
        $ 50.6$                           &  % Bug1L
        $ 64.9$                           &  % Bug1R
        $ 25.5$                           &  % Bug2L
        $ 15.1$                           \\ % Bug2R
        \cline{2-12}
    & \NE &
        $\z5.7$\footnotesize{$\pm \z0.2$} &  % \BRAND
        $302.3$\footnotesize{$\pm  28.8$} &  % RandomWalk
        $279.7$\footnotesize{$\pm  20.1$} &  % WallBounce
        $249.3$\footnotesize{$\pm  60.8$} &  % WaveFront
        $\z6.1$                           &  % Bug0L
        \Failed                           &  % Bug0R
        $\z6.1$                           &  % Bug1L
        $ 73.4$                           &  % Bug1R
        $\z6.1$                           &  % Bug2L
        \Failed                           \\ % Bug2R
        \cline{2-12}
    & \SW &
        $ 11.1$\footnotesize{$\pm \z1.8$} &  % \BRAND
        $156.3$\footnotesize{$\pm  20.4$} &  % RandomWalk
        $147.2$\footnotesize{$\pm  13.2$} &  % WallBounce
        $171.3$\footnotesize{$\pm  58.7$} &  % WaveFront
        $\z7.5$                           &  % Bug0L
        $\z7.5$                           &  % Bug0R
        $\z7.5$                           &  % Bug1L
        $\z7.5$                           &  % Bug1R
        $\z7.5$                           &  % Bug2L
        $\z7.5$                           \\ % Bug2R
        \Xhline{2\arrayrulewidth}
    \multicolumn{12}{c}{\textbf{Office Scene ($13m \times 12.5m$)}} \\
    \Xhline{2\arrayrulewidth}
    % ========== DSIT ==========
    \NW & \SE &
        $ 22.8$\footnotesize{$\pm \z3.6$} &  % \BRAND
        $173.6$\footnotesize{$\pm  15.7$} &  % RandomWalk
        $148.0$\footnotesize{$\pm  10.6$} &  % WallBounce
        $ 48.0$\footnotesize{$\pm  22.2$} &  % WaveFront
        \Failed                           &  % Bug0L
        \Failed                           &  % Bug0R
        $ 88.4$                           &  % Bug1L
        $ 64.1$                           &  % Bug1R
        $ 25.4$                           &  % Bug2L
        $ 34.9$                           \\ % Bug2R
        \cline{1-12}
    \NE & \SW &
        $ 36.3$\footnotesize{$\pm \z5.7$} &  % \BRAND
        $243.3$\footnotesize{$\pm  19.8$} &  % RandomWalk
        $207.7$\footnotesize{$\pm  13.5$} &  % WallBounce
        $ 56.9$\footnotesize{$\pm  28.2$} &  % WaveFront
        \Failed                           &  % Bug0L
        $ 17.0$                           &  % Bug0R
        $ 66.6$                           &  % Bug1L
        $ 88.0$                           &  % Bug1R
        $ 23.9$                           &  % Bug2L
        $ 35.0$                           \\ % Bug2R
        \cline{1-12}
    \SW & \NE &
        $ 16.9$\footnotesize{$\pm \z4.3$} &  % \BRAND
        $263.9$\footnotesize{$\pm  24.4$} &  % RandomWalk
        $268.3$\footnotesize{$\pm  16.9$} &  % WallBounce
        $ 65.7$\footnotesize{$\pm  34.3$} &  % WaveFront
        \Failed                           &  % Bug0L
        $ 14.5$                           &  % Bug0R
        $ 39.2$                           &  % Bug1L
        $ 47.0$                           &  % Bug1R
        $ 36.0$                           &  % Bug2L
        $ 31.4$                           \\ % Bug2R
        \cline{1-12}
    \SE & \NW &
        $ 14.5$\footnotesize{$\pm \z0.7$} &  % \BRAND
        $266.3$\footnotesize{$\pm  19.2$} &  % RandomWalk
        $290.4$\footnotesize{$\pm  18.5$} &  % WallBounce
        $ 72.5$\footnotesize{$\pm  38.6$} &  % WaveFront
        $ 15.9$                           &  % Bug0L
        \Failed                           &  % Bug0R
        $ 47.4$                           &  % Bug1L
        $ 40.1$                           &  % Bug1R
        $ 37.7$                           &  % Bug2L
        $ 25.9$                           \\ % Bug2R
        \Xhline{2\arrayrulewidth}
    \\[-3mm]
    % \multicolumn{7}{l}{\small{
    %     \textsuperscript{\textdagger}
    %         All units are in meters. Failures are shown as red cross marks (\,\Failed\,).
    % }}
    \end{tabular}
    \label{tab:results}
    % \vspace{-5mm}
\end{table*}

\end{document}